\documentclass[letterpaper]{article} 
\usepackage{aaai2026}  
\usepackage{times}  
\usepackage{helvet}  
\usepackage{courier}  
\usepackage[hyphens]{url}  
\usepackage{graphicx} 
\usepackage{natbib}  
\usepackage{caption} 
\usepackage{algorithm}

\usepackage{enumitem}
\usepackage{multirow}
\usepackage{amsmath}
\usepackage{amsfonts}
\usepackage{graphicx}
\usepackage{subcaption}
\usepackage{booktabs}

\usepackage{algpseudocode}
\usepackage{xcolor}

\usepackage{newfloat}
\usepackage{listings}
\DeclareCaptionStyle{ruled}{labelfont=normalfont,labelsep=colon,strut=off} 
\floatstyle{ruled}
\newfloat{listing}{tb}{lst}{}
\floatname{listing}{Listing}
\title{Lifelong Domain Adaptive 3D Human Pose Estimation}
\author{
    Qucheng Peng\textsuperscript{\rm 1}, Hongfei Xue\textsuperscript{\rm 2}, 
    Pu Wang\textsuperscript{\rm 2},
    Chen Chen\textsuperscript{\rm 1}\\
}
\affiliations{
    \textsuperscript{\rm 1}Center for Research in Computer Vision, University of Central Florida, Orlando, Florida, USA\\

    \textsuperscript{\rm 2}Department of Computer Science, University of North Carolina at Charlotte, Charlotte, North Carolina, USA\\
    qucheng.peng@ucf.edu, hongfei.xue@charlotte.edu, pu.wang@charlotte.edu,
    chen.chen@crcv.ucf.edu\\
}

\usepackage{bibentry}

\begin{document}

\maketitle

\begin{abstract}
3D Human Pose Estimation (3D HPE) is vital in various applications, from person re-identification and action recognition to virtual reality. However, the reliance on annotated 3D data collected in controlled environments poses challenges for generalization to diverse in-the-wild scenarios. Existing domain adaptation (DA) paradigms like general DA and source-free DA for 3D HPE overlook the issues of non-stationary target pose datasets. To address these challenges, we propose a novel task named lifelong domain adaptive 3D HPE. \emph{To our knowledge, we are the first to introduce the lifelong domain adaptation to the 3D HPE task.}  In this lifelong DA setting, the pose estimator is pretrained on the source domain and subsequently adapted to distinct target domains. Moreover, during adaptation to the current target domain, the pose estimator cannot access the source and all the previous target domains. The lifelong DA for 3D HPE involves overcoming challenges in adapting to current domain poses and preserving knowledge from previous domains, particularly combating catastrophic forgetting. We present an innovative Generative Adversarial Network (GAN) framework, which incorporates 3D pose generators, a 2D pose discriminator, and a 3D pose estimator. This framework effectively mitigates domain shifts and aligns original and augmented poses. Moreover, we construct a novel 3D pose generator paradigm, integrating pose-aware, temporal-aware, and domain-aware knowledge to enhance the current domain's adaptation and alleviate catastrophic forgetting on previous domains. Our method demonstrates superior performance through extensive experiments on diverse domain adaptive 3D HPE datasets. Code available at \url{https://github.com/davidpengucf/lifelongpose}.
\end{abstract}

\section{Introduction}
\label{sec:intro}

3D Human Pose Estimation (3D HPE) involves predicting the 3D coordinates of human joints from images or videos, providing a crucial foundation for applications such as person re-identification \cite{su2017pose}, action recognition \cite{lu2023hard, yan2023mae,peng2025navigscene}, virtual reality \cite{guzov2021human, yi2023mime,peng3d}. 
The 2D-to-3D lifting paradigm \cite{pavllo20193d,zheng20213d,zhao2023poseformerv2,peng2024dual}, which predicts 3D poses based on 2D poses \cite{peng2023source,peng2025exploiting}, stands as the most widely adopted pipeline in 3D HPE. Despite its significance, annotated 3D data are typically collected in controlled laboratory settings, featuring indoor environments and a limited range of actions performed by a few individuals. Consequently, pose estimators trained on such labeled datasets encounter difficulties in generalizing to diverse in-the-wild scenarios. Thus, the concept of Domain Adaptation (DA) for 3D HPE \cite{gholami2022adaptpose, chai2023poseda, liu2023posynda} becomes imperative, aiming to integrate knowledge from labeled (\textbf{source}) data into a pose estimator capable of effective generalization on unlabeled (\textbf{target}) data. 

Existing adaptation settings in 3D HPE fail to account for the evolving nature of pose distributions. While source-free domain adaptation methods \cite{guan2022DynBOA,nam2023CycleAdapt} enable co-training with all target poses, they assume static distributions and do not address realistic distribution shifts. In practice, target pose distributions are inherently non-stationary—\emph{they continuously evolve due to changing environments and the natural variability of individuals performing actions.} This is particularly evident in autonomous driving scenarios, where pose estimators must adapt across diverse contexts: predicting pedestrian intentions in outdoor environments, monitoring passenger safety inside vehicles.

\begin{figure*}[!ht]
  \centering   \includegraphics[width=1.0\linewidth]{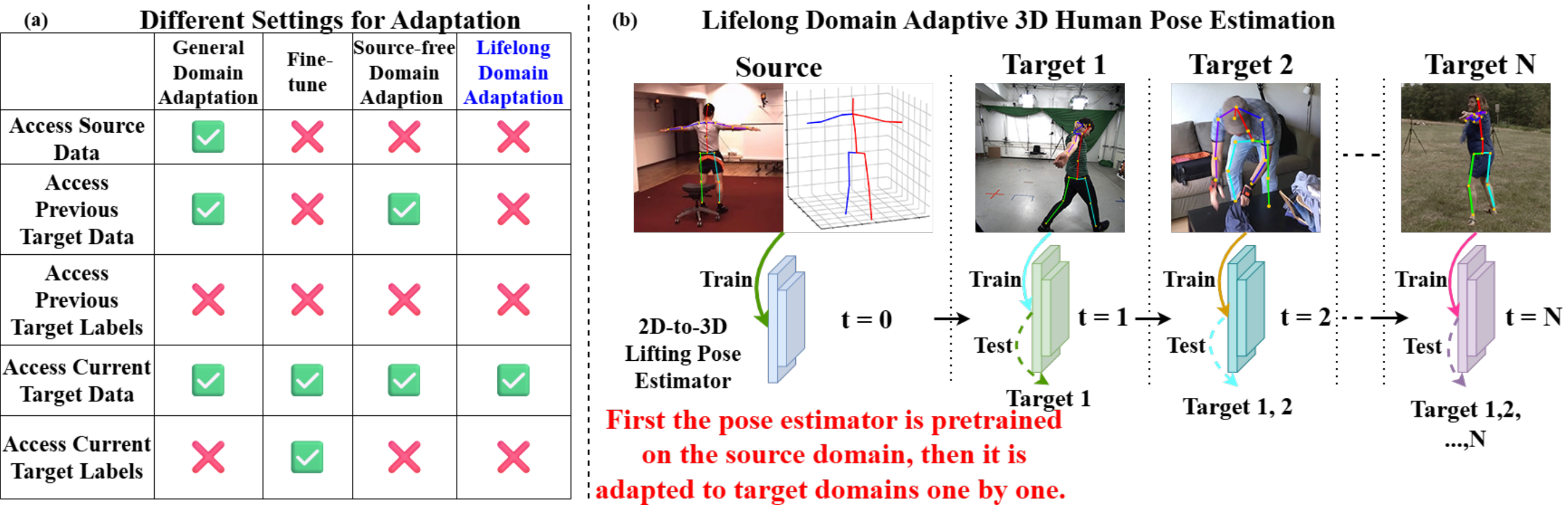}
   \caption{(a) Comparisons among general domain adaptation \cite{generalDA}, fine-tune \cite{finetune}, source-free/test-time domain adaptation \cite{testtimeDA}, and lifelong domain adaptation \cite{lifelongDA}. (b) The paradigm of lifelong domain adaptive 3D human pose estimation. }
   \label{fig:problem}
\end{figure*}

To address these challenges, we propose a novel task: \textbf{lifelong domain adaptive 3D human pose estimation}, as shown in Fig. \ref{fig:problem}b. \emph{To our knowledge, this is the first time a lifelong DA setting has been introduced for the 3D HPE task.} Unlike the adaptation settings presented in Fig. \ref{fig:problem}a, the lifelong approach begins with pretraining a 2D-to-3D pose estimator on 2D-3D pose pairs from the source domain. The model is then sequentially adapted to distinct target domains, one at a time, without access to annotations. During each adaptation phase, the pose estimator cannot reference poses from the source or any previously encountered target domains. Our objective is to develop an estimator that performs effectively across both the current and all previously encountered target domains. \emph{The lifelong adaptation differs from source-free/test-time adaptation, as it does not retain access to data from previous target domains.}

The proposed lifelong domain adaptation framework for 3D HPE addresses two main challenges: adapting the 2D-to-3D lifting pose estimator to new domains and preserving knowledge from previous domains. 
To counter these issues, we introduce a framework featuring 3D pose generators, a 2D pose discriminator, and a 2D-to-3D lifting pose estimator, employing a generative adversarial network (GAN) \cite{goodfellow2014generative} structure to minimize domain shifts. This framework ensures high-quality adaptations by aligning 2D and 3D poses and incorporates a novel 3D pose generator that utilizes pose-aware, temporal-aware, and domain-aware information to enhance adaptation and mitigate catastrophic forgetting. Additionally, a 2D pose diffusion sampler is implemented for efficient domain-aware prior generation. Our contributions can be summarized in three main aspects:

\begin{itemize}[noitemsep,leftmargin=*]
\item We introduce lifelong domain adaptive 3D HPE, addressing sequential domain shifts without access to previous domain data. \emph{This is the first work to tackle lifelong domain adaptation in 3D HPE}.
\item We demonstrate that mitigating catastrophic forgetting requires synergistic integration of 3D generators, adversarial 2D alignment, and exponential moving average n.
\item We conduct comprehensive experiments across multiple benchmarks, demonstrating our approach significantly outperforms existing methods in lifelong adaptation.
\end{itemize}

\begin{figure}[!ht]
  \centering
  \includegraphics[width=1.0\linewidth]{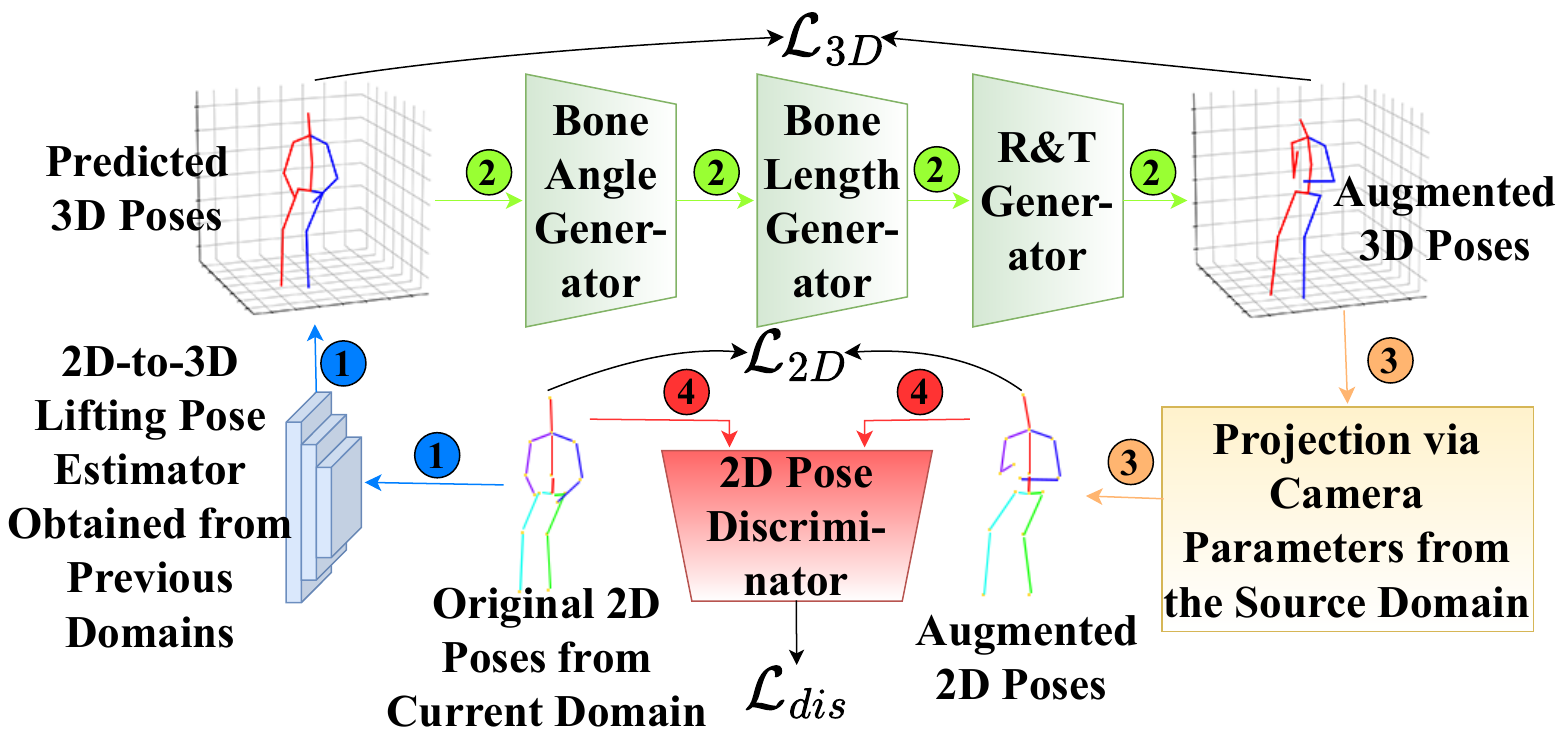}
  \caption{Overall adaption framework of our proposed lifelong domain adaptive 3D HPE approach at time $t = j$.}
  \label{fig:model}
\end{figure}

\begin{figure*}[!ht]
  \centering
  \includegraphics[width=0.9\linewidth]{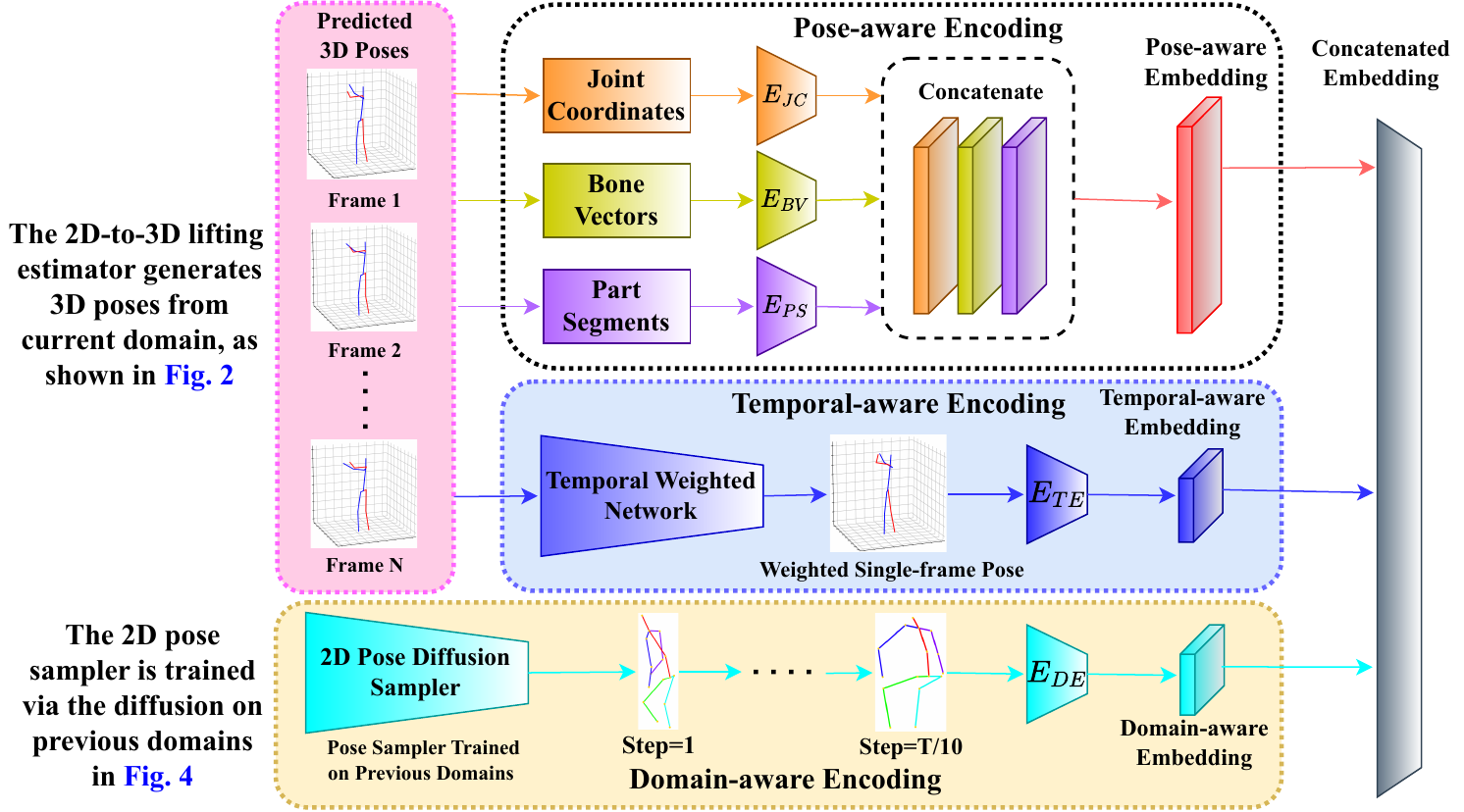}
  \caption{Details of the unified paradigm for each of the three 3D pose generators in Fig. \ref{fig:model}. In contrast to existing generators, we introduce part segments (in purple color) for improved pose-aware encoding and introduce additional temporal-aware encoding (in blue color), leading to better adaptation on the current domain. Besides, we employ an unconditional 2D pose diffusion sampler to generate the domain-aware embedding (in cyan color), effectively mitigating catastrophic forgetting.}
  \label{fig:frame}
\end{figure*}

\begin{figure*}[!htb]
  \centering
  \includegraphics[width=1.0\linewidth]{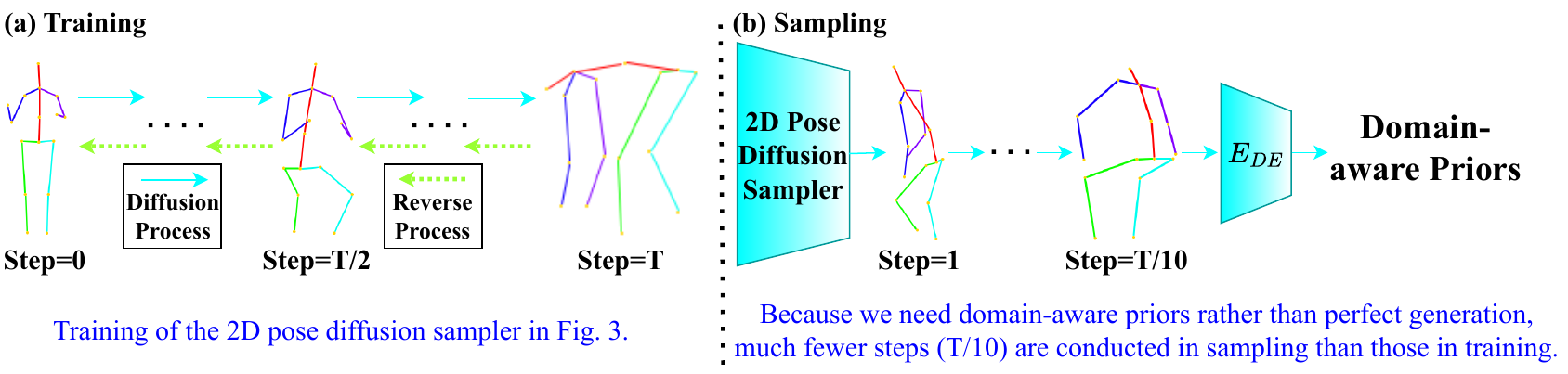}
  \caption{Training (a) and sampling (b) processes of the 2D pose diffusion sampler.}
  \label{fig:diffu}
\end{figure*}

\section{Related Work}
\label{sec:related}
\noindent\textbf{3D Human Pose Estimation.} The 2D-to-3D lifting paradigm dominates 3D HPE, where 2D pose estimators \cite{li2020cascaded,wang2020deep} generate predictions that are then transformed into 3D poses. Key approaches include dilated temporal convolutions \cite{pavllo20193d}, transformer-based methods \cite{zheng20213d}, mixed sequence-to-sequence encoders \cite{zhang2022mixste}, and frequency domain techniques \cite{zhao2023poseformerv2}.

\noindent\textbf{Lifelong Domain Adaptation.} Existing methods focus primarily on classification tasks. TENT \cite{wang2020tent} minimizes generalization errors through entropy minimization, while CoTTA \cite{wang2022continual} addresses error accumulation using weight-averaged predictions and stochastic restoration. Other works explore gradual shifts \cite{marsden2022gradual}, symmetric measurements \cite{dobler2023robust}, and extensions to object detection \cite{yang2022detect} and person re-identification \cite{huang2022reid}. \emph{We present the first lifelong domain adaptation approach for 3D human pose estimation}.

\noindent\textbf{Domain Adaptive 3D Human Pose Estimation.} Current DA methods for 3D HPE use both labeled source and unlabeled target data during training. Approaches include GANs for domain discrimination \cite{gholami2022adaptpose}, global-local alignment strategies \cite{chai2023poseda}, multi-hypotheses networks with source augmentation \cite{liu2023posynda}, and hybrid optimization-learning methods \cite{jiang2024back}. Source-free adaptations combine poses and 3D human shapes \cite{guan2022DynBOA,nam2023CycleAdapt}. \emph{Our work focuses on lifelong domain adaptive 3D HPE, where the estimator processes one pose dataset at a time rather than accessing multiple domains simultaneously}.

\section{Methodology}
\label{sec:method}

\begin{table*}[tb]
    \centering
    \resizebox{1.0\linewidth}{!}{
    \begin{tabular}{c  c  c  c  c  c  c}
         \toprule
          Time & Method & S5 & S6 & S7 & S8 & Avg\\
         \hline
         t = 0 & Source-only  & 53.5/46.3 & 56.3/47.7 & 50.4/41.9 & 46.6/35.1 & - \\
         \hline
         \multirow{6}{*}{t = 4} & {Adaptpose-LL \cite{gholami2022adaptpose}} &  52.8/46.3 & 54.3/46.0 & 47.5/40.9 & 42.1/30.0 & 49.2/40.8\\
          & {RMT-Pose \cite{dobler2023robust}} &  53.0/46.4 & 54.0/45.5 & 47.1/40.8 & 40.9/28.7 & 48.8/40.4\\
         & {CoTTA-Pose \cite{wang2022continual}} &  52.2/45.7 & 53.9/45.1 & 46.7/40.3 & 42.0/29.8 & 48.7/40.2\\
         & {CycleAdapt-LL} \cite{nam2023CycleAdapt} &   52.5/46.0 & 54.0/45.5 & 46.9/40.4 & 41.7/30.1 & 48.8/40.5\\
         & {PoseDA-LL \cite{chai2023poseda}} &  51.5/44.9 & 51.9/44.5 & 46.2/39.5 & 40.9/28.6 &47.6/39.4\\
         & {Ours} & \textbf{48.7/42.5} & \textbf{48.6/40.8} & \textbf{42.3/36.9} & \textbf{40.0/27.4} & \textbf{44.9/36.9}\\ 

         \bottomrule
    \end{tabular}}
\caption{Cross-scenario adaptation on H3.6M: S1 $\rightarrow$ S5, S6, S7, S8. MPJPE ($\downarrow$)/PA-MPJPE ($\downarrow$) as metrics.}
\label{tab:h36m}
\end{table*}

\begin{table*}[tb]
    \centering
    \resizebox{1.0\linewidth}{!}{
    \begin{tabular}{c c c c c c c c c}
         \toprule
          Time & Method & TS1 & TS2 & TS3 & TS4 & TS5 & TS6 & Avg \\
         \hline
         t = 0 & Source-only & 83.0/59.2 & 105.4/74.5 & 90.9/60.6 & 104.7/70.9 & 109.4/64.5 & 102.7/77.4 & - \\
         \hline
         \multirow{6}{*}{t = 6} & {Adaptpose-LL} & 69.0/50.1 & 83.8/55.7 & 69.4/46.5 & 82.7/58.2 & 90.0/59.4 & 98.5/66.8 & 82.2/56.1 \\
         & {RMT-Pose} & 68.8/49.7 & 83.5/55.2 & 69.0/46.3 & 82.5/57.7 & 88.6/57.7 & 97.0/66.2 &  81.6/55.5\\
         & {CoTTA-Pose}  & 68.4/49.5 & 83.1/54.9 & 68.5/45.8 & 81.9/57.2 & 89.0/58.5 & 96.6/65.5 &  81.3/55.2\\
         & {CycleAdapt-LL} & 68.6/49.7 & 83.5/55.1 & 68.7/46.2 & 82.0/57.3 & 89.4/58.6 & 96.4/65.4 & 81.4/55.4\\
         & {PoseDA-LL} & 67.8/48.5 & 82.4/54.1 & 67.9/44.8 & 81.3/56.4 & 88.5/57.6 & 96.0/65.3 &  80.7/54.5\\
         & {Ours} & \textbf{61.1/42.9} & \textbf{74.9/49.7} & \textbf{62.3/40.9} & \textbf{75.4/52.3} & \textbf{83.9/55.5} & \textbf{94.3/62.6} & \textbf{75.3/50.7} \\

         \bottomrule
    \end{tabular}}
    \caption{Cross-dataset adaptation on H3.6M $\rightarrow$ 3DHP: TS1,...,TS6. MPJPE ($\downarrow$)/PA-MPJPE ($\downarrow$) as metrics.}
    \label{tab:3dhp}
\end{table*}

\begin{table*}
\centering
  \resizebox{0.8\linewidth}{!}{
  \begin{tabular}{c c | c  c  c | c  c  c}
    \hline
     &
      & \multicolumn{3}{c|}{H3.6M $\rightarrow$ 3DHP $\rightarrow$ 3DPW} &
      \multicolumn{3}{c}{H3.6M $\rightarrow$ 3DPW $\rightarrow$ 3DHP} \\
      \hline
      
    Time & Method & 3DHP & 3DPW & Avg & 3DHP & 3DPW & Avg \\
    \hline
    t = 0 & Source-only & 96.4/66.5 & 103.3/63.6 & - & 96.4/66.5 & 103.3/63.6 & -\\
    \hline
    \multirow{6}{*}{t = 2} & {Adaptpose-LL} & 90.5/64.1 & 88.2/50.2 & 89.4/57.2 & 80.5/53.4 & 97.6/62.9 & 89.1/58.2\\
         & {RMT-Pose} & 90.9/64.3 & 88.5/50.4 & 89.7/57.4 & 79.9/53.3 & 95.0/61.5 & 87.5/57.4\\
         & {CoTTA-Pose} & 90.0/63.8 & 88.7/50.5 & 89.4/57.2 & 81.1/53.6 & 93.2/60.0 & 87.2/56.8 \\
         & {CycleAdapt-LL} & 90.2/64.1 & 88.3/50.2 & 89.3/57.2 & 80.0/53.3 & 95.9/62.3 & 88.0/57.8\\
         & {PoseDA-LL} & 88.9/62.1 & 87.6/49.4 & 88.3/55.8 & 79.8/53.0 & 91.5/53.8 & 85.7/53.4\\
         & {Ours} & \textbf{75.3/51.1} & \textbf{81.7/45.6} & \textbf{78.5/48.4} & \textbf{78.3/52.2} & \textbf{83.7/46.9} & \textbf{81.0/49.6}\\

    \bottomrule
  \end{tabular}}
  \caption{Multi-dataset adaptation on ``H3.6M (Source) $\rightarrow$3DHP$\rightarrow$3DPW" and ``H3.6M (Source) $\rightarrow$3DPW$\rightarrow$3DHP". Values are MPJPE ($\downarrow$)/PA-MPJPE ($\downarrow$).}
  \label{tab:multi}
\end{table*}

\begin{table}[!ht]
  \begin{subtable}{1.0\linewidth}
  \resizebox{1.0\linewidth}{!}{
  \begin{tabular}{ c | c  c | c  c}
    \hline
      & \multicolumn{2}{c|}{H3.6M $\rightarrow$ 3DHP $\rightarrow$ 3DPW} &
      \multicolumn{2}{c}{H3.6M $\rightarrow$ 3DPW $\rightarrow$ 3DHP} \\
      \hline
       Method & 3DHP & 3DPW & 3DHP & 3DPW \\
    \hline
     {Ours w/o PS} & 80.2/56.1 & 85.2/48.9 & 79.6/52.9 & 86.8/49.3\\
     {Ours w/o TE} & 79.6/55.8 & 83.3/47.4 & 79.0/52.7 & 87.1/50.0\\
     {Ours w/o DE} & 83.5/57.4 & 83.7/47.6 & 79.2/52.7 & 88.2/51.7\\
     {Ours} & \textbf{75.3/51.1} & \textbf{81.7/45.6} & \textbf{78.3/52.2} & \textbf{83.7/46.9}\\

    \bottomrule
  \end{tabular}}
    \caption{Ablation of 3D pose generators.}
    \label{tab:ab-gene}
  \end{subtable}
  \hfill
  \begin{subtable}{1.0\linewidth}
  \resizebox{1.0\linewidth}{!}{
  \begin{tabular}{ c | c  c | c  c}
    \hline
      & \multicolumn{2}{c|}{H3.6M $\rightarrow$ 3DHP $\rightarrow$ 3DPW} &
      \multicolumn{2}{c}{H3.6M $\rightarrow$ 3DPW $\rightarrow$ 3DHP} \\
      \hline
      
     Method & 3DHP & 3DPW & 3DHP & 3DPW \\
    \hline
     {Ours w/o $\mathcal{L}_{2D}$} & 78.4/54.1 & 83.9/48.0 & 80.8/53.5 & 86.2/48.7\\
     {Ours w/o $\mathcal{L}_{3D}$} & 80.7/56.4 & 85.4/49.2 & 82.1/54.0 & 87.6/49.3\\
     {Ours w/o $\mathcal{L}_{dis}$} & 82.5/58.3 & 85.8/49.4 & 83.9/57.6 & 88.0/51.2\\
     {Ours w/o EMA } & 81.2/57.6 & 83.7/47.8 & 80.4/52.9 & 88.5/51.8\\
     {Ours} & \textbf{75.3/51.1} & \textbf{81.7/45.6} & \textbf{78.3/52.2} & \textbf{83.7/46.9}\\

    \bottomrule
  \end{tabular}}
  \caption{Ablation of overall framework.}
  \label{tab:ab-frame}
  \end{subtable}
\caption{Ablation study of (a) 3D pose generators and (b) the overall framework when $t = 2$ for the two multi-dataset adaptation tasks  ``H3.6M$\rightarrow$3DHP$\rightarrow$3DPW" and ``H3.6M$\rightarrow$3DPW$\rightarrow$3DHP". }
\end{table}

\noindent\textbf{2D-to-3D Lifting HPE.} The predominant paradigm in 3D Human Pose Estimation (3D HPE) involves 2D-to-3D lifting \cite{pavllo20193d,zheng20213d,zhang2022mixste,zhao2019semgcn}. This paradigm assumes that $x^{sr}_{i} \in \mathbb{R}^{J \times 2}$ represents the 2D coordinates of $J$ keypoints in a sample from the labeled source domain (with 2D poses as input). Correspondingly, $y^{sr}_{i} \in \mathbb{R}^{J \times 3}$ denotes the corresponding 3D positions in the camera coordinate system (with 3D poses as output). The source domain, denoted as $sr = \{(x_{i}^{sr}, y_{i}^{sr})\}_{i=1}^{M_{sr}}$, contain $M_{sr}$ 2D-3D pose pairs. The 2D-to-3D lifting pose estimator, denoted as $\mathcal{P}: x^{sr}_{i} \mapsto \hat{y}_{i}^{sr}$, predicts the corresponding 3D pose positions $\hat{y}_{i}^{sr}$. However, the fully-supervised paradigm \cite{pavllo20193d,zheng20213d} optimized for source poses is inadequate for addressing the DA problem due to the lack of considering domain shifts between source and target domains.

\noindent\textbf{Problem Statement.} In our new lifelong setting for 3D HPE, the pose estimator $\mathcal{P}$ is initially pre-trained on the labeled source domain $sr$. \emph{We define the timestamp for training on source poses as $t = 0$.} Subsequently, it undergoes adaptation to target domains $tg_1,..,tg_N$ sequentially, as shown in Fig. \ref{fig:problem}(b). For the training at time $t = j \in [1,N]$, $\mathcal{P}$ is exclusively exposed to the target domain $tg_j = \{x_{i}^{tg_j}\}_{i=1}^{M_j}$. During testing at time $t = j$, the evaluation focuses not only on performance for the current target domain $tg_j$ but also on performance across all previous target domains $tg_1,..,tg_{j-1}$. This approach closely mirrors the handling of non-stationary data in real-world scenarios. However, within the lifelong learning setting, addressing catastrophic forgetting of previous target domains becomes a challenge.

\noindent\textbf{Overview of the Proposed Method.} In Fig. \ref{fig:model}, we present the comprehensive pipeline of the lifelong 3D HPE for DA at time $t = j$. Given a 2D-to-3D lifting pose estimator $\mathcal{P}$ that has finished the adaptation on previous domains $tg_1,tg_2,..,tg_{j-1}$, it is initially utilized to predict pseudo 3D poses based on 2D poses from the current domain $tg_j$. Subsequently, the estimated 3D poses undergo augmentation through three distinctive 3D pose generators, each designed to encode pose-aware, temporal-aware, and domain-aware knowledge, as shown in Fig. \ref{fig:frame}.

The augmented 3D poses are then projected onto augmented 2D poses using camera parameters from the source domain. The framework's update is facilitated through three distinct loss functions. One is $\mathcal{L}_{3D}$, which aligns predicted 3D poses with augmented 3D poses. Another one is $\mathcal{L}_{2D}$, which compares ground truth 2D poses with augmented 2D poses. Moreover, a 2D pose discriminator is introduced to distinguish between the two types of 2D poses. The min-max game among the 2D pose discriminator and the 3D pose generators is regulated by the third loss $\mathcal{L}_{dis}$. 

\noindent\textbf{3D Pose Generation.} In Fig. \ref{fig:frame}, we show the details of our proposed 3D pose generator paradigm illustrated in Fig. \ref{fig:model}. Consecutive 3D pose frames estimated by the 2D-to-3D lifting pose estimator serve as the foundation for constructing pose-aware, temporal-aware, and domain-aware embeddings. Projection neural networks $E_{JC},E_{BV},..,E_{DE}$  are utilized to project inputs such as joint coordinates or bone vectors to embeddings, and these embeddings are concatenated to form the input for 3D pose generators.

In prior works \cite{gholami2022adaptpose,chai2023poseda}, pose-aware encoding is achieved by extracting joint coordinates and bone vectors from 3D poses. However, these approaches overlook part-aware information, which is essential for a more comprehensive representation of the human body. Moreover, joints not physically connected or belonging to the same part in the human body model still exhibit relationships that warrant consideration. Consequently, we delineate six body part segments based on the human body: left hand, right hand, left leg, right leg, torso, and extended torso. The first five segments cover the five primary body parts, while the last segment --extended torso-- establishes connections for joints that lack physical linkage and do not belong to the same part. 

For temporal-aware encoding, multiple consecutive frames of 3D poses are input into a temporal weighted convolutional network, generating a weighted single-frame pose. This process encodes temporal-aware knowledge to enhance the 3D pose generator's synthesis capabilities.

For domain-aware encoding, we introduce domain-aware priors from previous domains as a mitigation strategy against catastrophic forgetting. We choose diffusion models \cite{ho2020denoising,song2020denoising} over GANs \cite{goodfellow2014generative} in terms of providing domain-aware priors due to the ability of diffusion models to preserve mode coverage and diversity, thus avoiding potential mode collapses in GANs \cite{xiao2021tackling}. This attribute is particularly crucial in the lifelong setting because there is a sequence of distribution shifts among the evolving target poses. Specifically, a 2D pose diffusion sampler, trained on 2D poses from prior domains via Denoising Diffusion Implicit Models (DDIM) \cite{song2020denoising}, is employed. The generated 2D poses are subsequently encoded as domain-aware embeddings. Fig. \ref{fig:diffu} illustrates the training and sampling process. 

In the training phase of the 2D pose diffusion sampler, the small size of pose data (typically (16,2) for 16 joints) compared to image size (typically (224, 224) for a standard ResNet \cite{he2016deep} input) allows for rapid convergence. During training, the maximum time step is set to $T$. Following \cite{ho2020denoising,song2020denoising}, we pick a random time step $k$ from Uniform[1, $T$]. During testing, where only domain-aware priors are required, a complete sampling with the same time step $T$ is unnecessary. Instead, we sample only $T/10$ steps, ensuring efficient processing. Consequently, we replace the randomly generated noise channels utilized in prior works \cite{gholami2022adaptpose,chai2023poseda} with more controllable and domain-aware priors for the generators. This substitution proves beneficial in mitigating catastrophic forgetting. 

By concatenating these diverse embeddings, we can construct more part-aware, temporal-aware, and domain-aware generators tailored for our lifelong DA tasks in 3D HPE.

\noindent\textbf{Optimization Process.} In this paragraph, we discuss the optimization process of our proposed method depicted in Fig. \ref{fig:model}. Assuming $t = j \in [1,N]$ (corresponding to the current domain $tg_{j}$), for a 2D pose $x_{i}^{tg_{j}} \in tg_{j}$, we derive the predicted 3D pose $\hat{y}_{i}^{tg_{j}} = \mathcal{P}_{j}(x_{i}^{tg_{j}})$ using the 2D-to-3D lifting pose estimator obtained before the initiation of $t = j$. The concatenation of the three generators is indicated as $G = G_{BA}\circ G_{BL}\circ G_{RT}$, where $G_{BA}$, $G_{BL}$, and $G_{RT}$ correspond to distinctive operations for bone angles, bone lengths, and rotation and translation. Consequently, the augmented 3D pose is expressed as $\tilde{y}_{i}^{tg_{j}} = G(\hat{y}_{i}^{tg_{j}})$, which is then subjected to projection via camera parameters from the source domain, resulting in the augmented 2D pose $\tilde{x}_{i}^{tg_{j}} = Proj(\tilde{y}_{i}^{tg_{j}})$. Subsequently, both the original 2D pose and the augmented 2D pose partake in the discrimination process through the 2D discriminator $D$.

We integrate the Mean Squared Error (MSE) loss along with the feedback loss \cite{gong2021poseaug,li2020pointaugment}, which ensures that the augmentation extent is sufficiently substantial for the 3D loss:
\begin{equation}
    \mathcal{L}_{3D}(x_{i}^{tg_{j}}) = \mathcal{L}_{MSE}(\hat{y}_{i}^{tg_{j}}, \tilde{y}_{i}^{tg_{j}}) + \lVert 1 - \exp|\hat{y}_{i}^{tg_{j}} - \tilde{y}_{i}^{tg_{j}}|_{\ell_1}\rVert.
\label{eq:3d}
\end{equation} The $\mathcal{L}_{3D}$ term ensures the similarity between predicted and augmented 3D poses within a reasonable range, allowing the augmented poses to differ from the predictions while adhering to human body constraints.

Regarding the 2D loss, we address the significant scale factor. Fully normalizing the scales of the original 2D pose and the augmented 2D pose could result in a loss of domain-aware knowledge. Conversely, entirely disregarding this factor would complicate the alignment between the two types of 2D poses. Hence, we propose the 2D loss:
\begin{equation}
    \mathcal{L}_{2D}(x_{i}^{tg_{j}}) = \mathcal{L}_{MSE}(x_{i}^{tg_{j}}, \tilde{x}_{i}^{tg_{j}}) + |\frac{x_{i}^{tg_{j}}}{\lVert x_{i}^{tg_{j}}\rVert} - \frac{\tilde{x}_{i}^{tg_{j}}}{\lVert \tilde{x}_{i}^{tg_{j}}\rVert}|_{\ell_1},
\label{eq:2d}
\end{equation} where the first term preserves the scales, and the second term normalizes the scales, striking a balance between the two considerations.

For the discrimination loss controlling the 3D generation process, we employ Wasserstein GANs with gradient penalties \cite{gulrajani2017improved} in our discrimination process:
\begin{multline}
    \mathcal{L}_{dis}(x_{i}^{tg_{j}}) = \mathbb{E}[D(x_{i}^{tg_{j}})] - \mathbb{E}[D(\tilde{x}_{i}^{tg_{j}})] \\
    + \alpha \mathbb{E}(1 - \lVert \nabla_{{k}_{i}^{tg_{j}}}D({k}_{i}^{tg_{j}})  \rVert),
\label{eq:dis}
\end{multline}
where ${k}_{i}^{tg_{j}} = \epsilon x_{i}^{tg_{j}} + (1-\epsilon)\tilde{x}_{i}^{tg_{j}}$, with $\epsilon$ randomly drawn from $U[0,1]$, and $\alpha$ serving as a trade-off parameter.

Based on the three proposed losses, the 3D generator $G$ is updated via:
\begin{equation}
    \mathcal{L}_{G}(x_{i}^{tg_{j}}) = \mathcal{L}_{3D}(x_{i}^{tg_{j}}) - \beta \mathcal{L}_{dis}(x_{i}^{tg_{j}}),
\label{eq:gen}    
\end{equation}
while both the 2D discriminator $D$ and the pose estimator $\mathcal{P}$ are optimized via:
\begin{equation}
    \mathcal{L}_{DP}(x_{i}^{tg_{j}}) = \mathcal{L}_{2D}(x_{i}^{tg_{j}}) + \gamma \mathcal{L}_{dis}(x_{i}^{tg_{j}}).
\label{eq:dp}
\end{equation}Here, $\beta$ and $\gamma$ are hyperparameters to balance the trade-off between different losses. Following optimization, an exponential moving average strategy (EMA) is applied to obtain the pose estimator $\mathcal{P}_{j+1}$ for the next time $t = j + 1$ as:
\begin{equation}
    \mathcal{P}_{j+1} = \eta \mathcal{P}_{j} + (1 - \eta)\mathcal{\hat{P}}_{j},
\label{eq:ema}
\end{equation}
where $\mathcal{P}_{j}$ is the model initialized before $t = j$ begins, and $\mathcal{\hat{P}}_{j}$ is the model updated after the adaptation on target domain $Tg_j$ is completed. The smoothing coefficient $\eta$ is set to 0.99. For the 3D pose generator $G$ and the 2D pose discriminator $D$, the parameters at timestamp $t = j + 1$ are inherited directly after the optimization at timestamp $t = j$.


\section{Experiments}
\label{sec:experiments}

\noindent\textbf{Datasets and Metrics.} Our approach is evaluated on three widely used 3D human pose datasets using a 16-keypoint body model and MPJPE/PA-MPJPE metrics. Human3.6M (H3.6M) features 7 indoor subjects (S1, S5-S8, S9, S11); we use S1 as source and S5-S8 as sequential targets. MPI-INF-3DHP (3DHP) contains indoor/outdoor scenes with six test sets (TS1-TS6) used for cross-dataset adaptation. 3DPW provides challenging in-the-wild scenes with 60 sequences. We evaluate on: (1) H3.6M cross-scenario adaptation S1$\rightarrow$S5$\rightarrow$S6$\rightarrow$S7$\rightarrow$S8, (2) cross-dataset adaptation H3.6M$\rightarrow$TS1$\rightarrow$...$\rightarrow$TS6, and (3) multi-dataset tasks H3.6M$\rightarrow$3DHP$\rightarrow$3DPW and H3.6M$\rightarrow$3DPW$\rightarrow$3DHP.

\noindent\textbf{Implementation Details.}
We use fully-connected layers for 3D pose generators and 2D pose estimator (VideoPose3D \cite{pavllo20193d}), and single convolutional layers for projection and temporal weighted networks. Learning rates are 1e-4 (generators/discriminator) and 5e-5 (pose estimator), with $\alpha=0.35$ and $\beta=\gamma=2.5$. We employ Adam optimizer \cite{kingma2014adam} for generators/discriminator and AdamW \cite{loshchilov2018decoupled} for the pose estimator. Training uses batch size 1024 with 27 frames \cite{gholami2022adaptpose,chai2023poseda}, 40 epochs for source pretraining, and 30 epochs per target domain adaptation.

For the 2D diffusion pose sampler, we use U-Net \cite{ronneberger2015unet} with batch size 64, Adam optimizer (lr=2e-4), and 10 training epochs with sampling steps in Uniform[1, 400]. \emph{We use 40 sampling steps for pseudo 2D pose generation, prioritizing domain-aware priors over perfect reconstruction for efficiency.}


\noindent\textbf{Baselines.} We establish lifelong DA baselines for 3D HPE by adapting existing methods from two categories. First, we extend 3D domain-adaptive HPE methods AdaptPose \cite{gholami2022adaptpose}, PoseDA \cite{chai2023poseda}, and CycleAdapt \cite{nam2023CycleAdapt} to lifelong settings as AdaptPose-LL, PoseDA-LL, and CycleAdapt-LL, transferring discriminations and augmentations between current and previous domain poses. Second, we adapt lifelong DA methods RMT \cite{dobler2023robust} and CoTTA \cite{wang2022continual} to 3D HPE as RMT-Pose and CoTTA-Pose, replacing classification losses with MSE losses.

\noindent\textbf{Quantitative Results.} Tables \ref{tab:h36m}-\ref{tab:multi} demonstrate our method's superior performance across all benchmarks. Our approach consistently outperforms PoseDA-LL \cite{chai2023poseda}, achieving average improvements of 2.7mm/2.5mm (MPJPE/PA-MPJPE) at $t=4$ (Table \ref{tab:h36m}) and 5.4mm/3.4mm at $t=6$ (Table \ref{tab:3dhp}). Notably, we achieve substantial gains of 6.7mm/5.6mm on TS1 at $t=6$. In the challenging multi-dataset adaptation scenario (Table \ref{tab:multi}), our method surpasses PoseDA-LL by 9.8mm/7.4mm on average, with particularly strong performance on 3DHP (13.6mm/11.0mm improvement at $t=2$). These quantitative results underscore our framework's effectiveness in both current domain adaptation and catastrophic forgetting mitigation across diverse lifelong learning scenarios. 

\begin{table*}[ht]
\centering
  \resizebox{0.7\linewidth}{!}{
  \begin{tabular}{c | c  c | c  c}
    \hline
      & \multicolumn{2}{c|}{H3.6M $\rightarrow$ 3DHP $\rightarrow$ 3DPW} &
      \multicolumn{2}{c}{H3.6M $\rightarrow$ 3DPW $\rightarrow$ 3DHP} \\
      \hline

    Sampling Steps & 3DHP & 3DPW & 3DHP & 3DPW \\
    \hline
     {step = 0 (Random Noise)} &  83.4/57.3 & 83.7/47.6 & 79.2/52.7 & 88.0/51.5\\
     {steps = T / 40} & 80.4/55.7 & 85.2/48.3 & 82.9/55.7 & 87.3/50.6 \\
     {steps = T / 20} & 76.8/52.6 & 82.2/46.1 & 79.8/53.5 & 84.2/47.4 \\
     {steps = T / 10 (Ours)} & \textbf{75.3/51.1} & \textbf{81.7/45.6} & \textbf{78.3/52.2} & \textbf{83.7/46.9}\\
     {steps = T / 5} & 77.0/53.3 & 82.8/46.5 & 80.6/54.1 & 85.0/47.9 \\
     {steps = T / 2} & 79.8/54.1 & 84.5/47.3 & 82.3/54.9 & 87.5/49.1 \\
     {steps = T} & 80.5/55.8 & 84.7/48.0 & 82.0/55.6 & 88.0/49.9 \\
    \bottomrule
  \end{tabular}}
  \caption{Analysis of the 2D pose sampler steps for the two multi-dataset adaptation tasks  H3.6M$\rightarrow$3DHP$\rightarrow$3DPW" and H3.6M$\rightarrow$3DPW$\rightarrow$3DHP" when $t = 2$. }
  \label{tab:ab-samp}
\end{table*}

\noindent\textbf{Ablation Study on 3D Pose Generators.} In Tab. \ref{tab:ab-gene}, we perform an ablation study on the components of 3D pose generators. 
Eliminating the joint coordinates and bone vectors employed in previous works \cite{gholami2022adaptpose,chai2023poseda}, our focus is on three key components: part segments (\textbf{PS}), temporal-aware embedding (\textbf{TE}), and domain-aware embedding (\textbf{DE}). Based on the results, we observe that \textbf{DE} emerges as the most crucial component for preserving knowledge from previous domains. The removal of \textbf{DE} results in a degradation of 8.2mm and 6.3mm on the MPJPE of 3DHP and 3DPW, respectively, based on the task ``H3.6M$\rightarrow$3DHP$\rightarrow$3DPW". 

\noindent\textbf{Ablation Study on the Overall Framework.} Tab. \ref{tab:ab-frame} evaluates each component: $\mathcal{L}_{2D}$, $\mathcal{L}_{3D}$, $\mathcal{L}_{dis}$, and EMA. EMA prevents catastrophic forgetting (5.9mm/6.5mm drops when removed), $\mathcal{L}_{2D}$/$\mathcal{L}_{3D}$ enable current domain adaptation, and $\mathcal{L}_{dis}$ preserves knowledge while adapting (7.2mm/4.1mm increases when removed). All components are essential.

\noindent\textbf{Analysis of 2D Pose Sampler's Sampling Steps.} Due to the lifelong setting, it is necessary to incorporate knowledge from previously learned domains, and that is why we propose the 2D pose sampler to generate domain-aware priors. In such a case, it is meaningful to investigate how many steps of sampling are optimal for providing these priors. Based on the maximum training steps T$= 400$, we evaluate several values for sampling steps in Tab. \ref{tab:ab-samp}.


\begin{table*}[!ht]
\centering
\resizebox{0.8\linewidth}{!}{
\begin{tabular}{ c | c  c  c  c  c  }
  \hline
    Method &  S5 & S6 & S7 & S8 & Avg\\
  \hline
   {VAE} & 52.1/46.2 & 54.2/46.3 & 48.0/41.3 & 40.3/28.0 & 48.7/40.5\\
   {DDIM} & 51.4/45.6 & 53.6/45.9 & 46.2/39.8 & 40.3/28.0 & 47.9/39.8\\
   {GAN (Ours)} & \textbf{48.7/42.5} & \textbf{48.6/40.8} & \textbf{42.3/36.9} & \textbf{40.0/27.4} & \textbf{44.9/36.9} \\
  \bottomrule
\end{tabular}}
\caption{Comparisons of generative models in 3D pose generation on H3.6M: S1 $\rightarrow$ S5, S6, S7, S8 when $t = 4$}
\label{tab:ab-gen-h36m}
\end{table*}

\begin{table*}[!ht]
\centering
\resizebox{1.0\linewidth}{!}{
\begin{tabular}{ c | c  c  c  c  c  c  c}
  \hline
    Method &  TS1 & TS2 & TS3 & TS4 & TS5 & TS6 & Avg  \\
  \hline
   {VAE} & 69.8/50.4 & 84.0/55.1 & 69.7/48.1 & 81.9/57.4 & 87.8/56.6 & 95.3/64.8 & 81.4/55.4\\
   {DDIM} & 67.1/48.8 & 82.6/54.5 & 67.7/47.4 & 79.2/55.8 & 86.4/55.9 & 95.0/63.3 & 79.7/54.3\\
   {GAN (Ours)} & \textbf{61.1/42.9} & \textbf{74.9/49.7} & \textbf{62.3/40.9} & \textbf{75.4/52.3} & \textbf{83.9/55.5} & \textbf{94.3/62.6} & \textbf{75.3/50.7} \\
  \bottomrule
\end{tabular}}
\caption{Comparisons of generative models in 3D pose generation on H3.6M $\rightarrow$ 3DHP: TS1, TS2,..., TS6 when $t = 6$}
\label{tab:ab-gen-3dhp}
\end{table*}

\noindent\textbf{Analysis of 3D Pose Generation Method.} In this paper, we utilize GAN \cite{goodfellow2014generative} for the interpretable generation of 3D poses. In Table \ref{tab:ab-gen-3dhp} and Table \ref{tab:ab-gen-h36m}, we compare our stage-by-stage generation approach with one-stage generative methods such as VAE \cite{kingma2019vae} and DDIM \cite{song2020denoising}. The results highlight the superiority of GAN over other generative models in the context of lifelong domain adaptation for 3D HPE.

\begin{figure*}[!htb]
  \centering
  \includegraphics[width=1.0\linewidth]{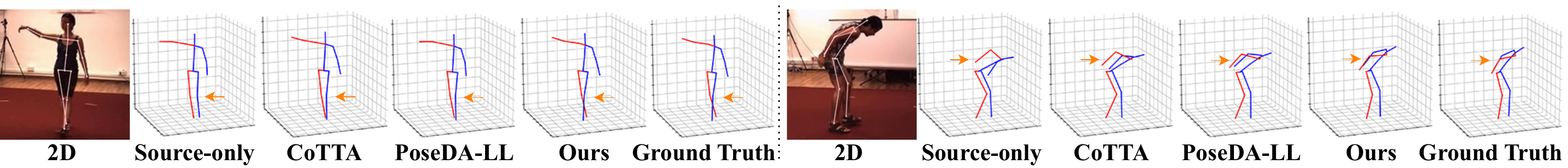}
  \caption{Visualization of H3.6M based on H3.6M: S1 $\rightarrow$ S5, S6, S7, S8 in Tab. \ref{tab:h36m}. The results are generated via the pose estimator obtained after $t = 4$. Left side is from S5 and right side is from S6.}
  \label{fig:h36m}
\end{figure*}

\begin{figure*}[!htb]
  \centering
  \includegraphics[width=1.0\linewidth]{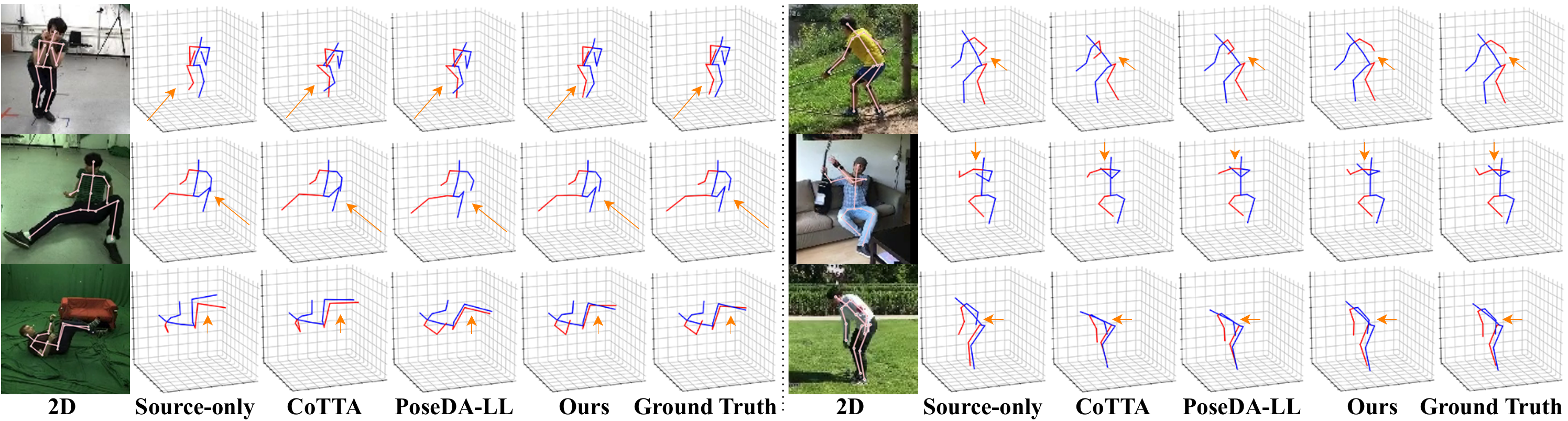}
  \caption{Visualization of 3DHP (left) in Tab. \ref{tab:3dhp} based on H3.6M $\rightarrow$ 3DHP: TS1,...,TS6 after $t = 6$, and 3DPW (right) based on H3.6M$\rightarrow$3DPW$\rightarrow$3DHP in Tab. \ref{tab:multi} generated after $t = 2$.}
  \label{fig:hppw}
\end{figure*}

\noindent\textbf{Qualitative Results.} Qualitative results are depicted in Fig. \ref{fig:h36m} and Fig. \ref{fig:hppw}. We include Source-only, CoTTA-Pose, PoseDA-LL, Ours, and Ground Truth for qualitative comparisons. 
It is evident from the visual comparisons that our method outperforms other baselines significantly.

\section{Conclusion}
\label{sec:conclusion}

In this study, we propose lifelong domain adaptive 3D human pose estimation to address non-stationary target datasets in real-world scenarios. Our framework comprises 3D pose generators, a 2D pose discriminator, and a pose estimator that leverage a GAN structure to mitigate domain shifts through a min-max game. We introduce a novel 3D pose generator paradigm incorporating pose-aware, temporal-aware, and domain-aware encoding, where pose and temporal components enhance current domain adaptation while the domain-aware component mitigates catastrophic forgetting. Extensive experiments demonstrate significant performance advantages over existing approaches.

\bibliography{aaai2026}

\section{Overview}

The supplementary material is organized into the following sections:

\begin{itemize}
    \item Additional quantitative results.
    \item Additional qualitative results.
    \item Additional ablation study.
    \item Analysis of varied 2D poses.
    \item Analysis of varied 2D-lifting-3D backbones.
    \item Analysis of the number of frames.
    \item Analysis of domain-aware priors' generation.
    \item Hyperparameters analysis.
    \item Details of part segments.
    \item Explanations of the three generators.
    \item Details of domain-aware encoding.
    \item Computational complexity and runtime analysis.
    \item Broader impacts.
   
\end{itemize}

\section{Extra Quantitative Results}
\label{sec:quant}

We present the quantitative results in an order deviating from the data sequence outlined in the main paper, to illustrate the robustness of our proposed method across varied data arrival patterns. Contrary to the data sequences detailed in the main paper, such as "S1 (Source)$\rightarrow$ S5$\rightarrow$S6$\rightarrow$S7$\rightarrow$S8" and "H3.6M (Source) $\rightarrow$TS1$\rightarrow$...$\rightarrow$TS6", we investigate alternative data sequences "S1 (Source) $\rightarrow$S8$\rightarrow$S7$\rightarrow$S6$\rightarrow$S5" and "H3.6M (Source) $\rightarrow$TS6$\rightarrow$...$\rightarrow$TS1" respectively to evaluate the method's effectiveness. Quantitative outcomes are reported at the final timestamp of H3.6M \cite{ionescu2013human3} in Tab. \ref{tab:h36m} when $t = 4$, and for 3DHP \cite{mehta2017monocular} in Tab. \ref{tab:3dhp} when $t = 6$.

\begin{table}[!ht]
\small
    \centering
    \caption{Cross-scenario adaptation on H3.6M: S1 $\rightarrow$ S8, S7, S6, S5. Dataset sequence is "S1 (Source) $\rightarrow$S8$\rightarrow$S7$\rightarrow$S6$\rightarrow$S5" at the final timestamp $t = 4$. Values are MPJPE ($\downarrow$)/PA-MPJPE ($\downarrow$).}
    \resizebox{1.0\linewidth}{!}{
    \begin{tabular}{c  c  c  c  c  c  c  c}
         \toprule
          Time & Method & Venue & S8 & S7 & S6 & S5 & Avg\\
         \hline
         t = 0 & Source-only &  & 46.6/35.1 & 50.4/41.9 & 56.3/47.7 & 53.5/46.3 & - \\
         \hline
         \multirow{5}{*}{t = 4} & {Adaptpose-LL \cite{gholami2022adaptpose}} & CVPR'22 & 46.9/35.5 & 49.6/42.2 & 52.4/43.8 & 50.5/43.7 & 49.9/41.3 \\
          & {RMT-Pose \cite{dobler2023robust}} & CVPR'23 & 46.8/34.7 & 49.3/42.1 & 50.8/42.6 & 49.7/42.9 & 49.2/40.6  \\
         & {CoTTA-Pose \cite{wang2022continual}} & CVPR'22 & 46.3/34.1 & 48.8/41.9 & 48.8/41.3 & 48.5/42.6 & 48.1/40.0\\
         & {PoseDA-LL \cite{chai2023poseda}} & ICCV'23 & 45.9/34.0 & 48.4/41.5 & 48.5/41.3 & 47.9/42.0 & 47.7/39.7\\
         & {Ours} & & \textbf{41.5/29.3} & \textbf{43.3/36.8} & \textbf{48.0/40.2} & \textbf{46.7/36.8} & \textbf{44.9/35.8}\\ 

         \bottomrule
    \end{tabular}}
    \label{tab:h36m-sm}
\end{table}

\begin{table}[!ht]
    \centering
    \caption{Cross-dataset adaptation on H3.6M $\rightarrow$ 3DHP: TS6,TS5,...,TS1. Dataset sequence is "H3.6M (Source) $\rightarrow$TS6$\rightarrow$...$\rightarrow$TS1" at the final timestamp $t = 6$. Values are MPJPE ($\downarrow$)/PA-MPJPE ($\downarrow$).}
    \resizebox{1.0\linewidth}{!}{
    \begin{tabular}{c c c c c c c c c}
         \toprule
          Time & Method & TS6 & TS5 & TS4 & TS3 & TS2 & TS1 & Avg \\
         \hline
         t = 0 & Source-only & 102.7/77.4 & 109.4/64.5 & 104.7/70.9 & 90.9/60.6 & 105.4/74.5 & 83.0/59.2 & - \\
         \hline
         \multirow{5}{*}{t = 6} & {Adaptpose-LL} & 100.5/72.9 & 103.5/63.0 & 84.4/60.8 & 65.8/42.7 & 78.8/51.4 & 61.1/42.7 & 82.4/55.6\\
         & {RMT-Pose} & 100.0/72.4 & 101.7/62.7 & 83.9/59.7 & 65.2/42.3 & 78.0/50.5 & 61.0/42.4 & 81.6/55.0\\
         & {CoTTA-Pose}  & 99.8/72.0 & 100.4/62.1 & 83.5/59.6 & 64.7/42.0 & 77.8/50.4 & 60.8/42.2 & 81.2/54.7\\
         & {PoseDA-LL} & 98.4/70.7 & 95.6/59.4 & 82.0/58.3 & 64.0/41.4 & 77.1/49.9 & 60.3/42.0 & 79.6/53.6\\
         & {Ours} & \textbf{95.0/64.3} & \textbf{86.2/56.1} & \textbf{76.9/54.2} & \textbf{61.1/39.9} & \textbf{74.5/48.2} & \textbf{57.8/40.5} & \textbf{75.4/50.5}\\

         \bottomrule
    \end{tabular}}
    \label{tab:3dhp-sm}
\end{table}

For instance, at $t = 4$ in Table \ref{tab:h36m-sm}, our method outperforms PoseDA-LL \cite{chai2023poseda} by an average of 2.8mm for MPJPE and 3.9mm for PA-MPJPE. Additionally, our method maintains a lead of 4.2mm for MPJPE and 3.4mm for PA-MPJPE on average at $t = 6$ in Table \ref{tab:3dhp-sm}. These quantitative results highlight the effectiveness and robustness of our proposed method in various data arrival scenarios.

\section{Extra Qualitative Results}
\label{sec:quali}

\begin{figure*}[tb]
  \centering
  \includegraphics[width=1.0\linewidth]{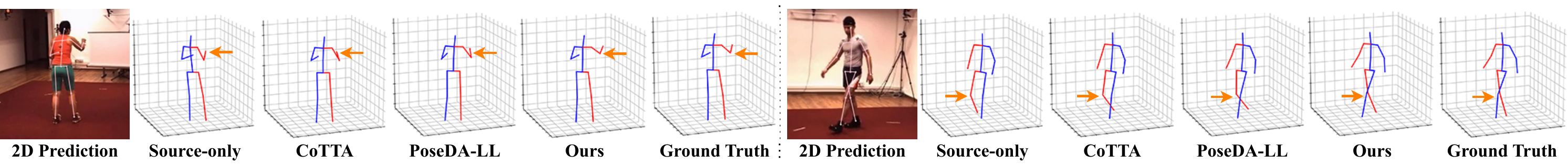}
  \caption{Visualization of H3.6M in Tab. \ref{tab:h36m}. The results are generated via pose estimator obtained after $t = 4$. Left side is from S7 and right side is from S8.}
  \label{fig:xh36m}
\end{figure*}

\begin{figure*}[tb]
  \centering
  \includegraphics[width=1.0\linewidth]{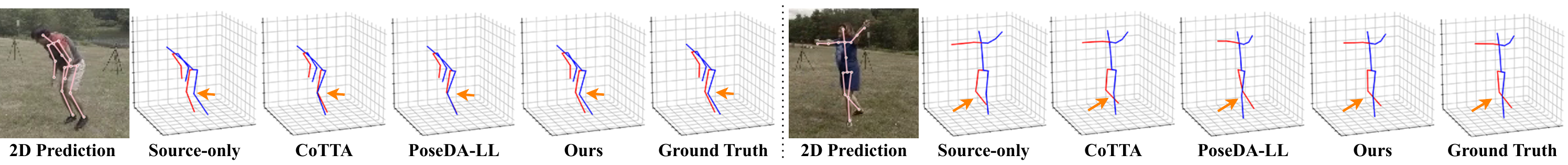}
  \caption{Visualization of 3DHP in Tab. \ref{tab:3dhp-sm}. The results are generated via pose estimator obtained after $t = 4$. Left side is from TS5 and right side is from TS6.}
  \label{fig:x3dhp}
\end{figure*}

\begin{figure*}[tb]
  \centering
  \includegraphics[width=1.0\linewidth]{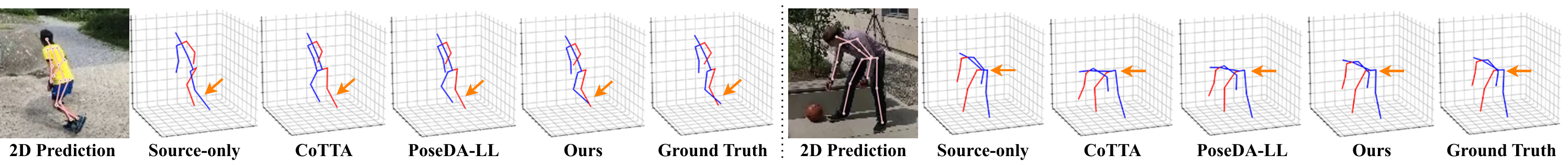}
  \caption{Visualization of 3DPW in the task "H3.6M (Source) $\rightarrow$ 3DPW $\rightarrow$ 3DHP". The results are generated via pose estimator obtained after $t = 2$.}
  \label{fig:x3dpw}
\end{figure*}

In this section, we provide additional qualitative results, which are presented in Fig. \ref{fig:xh36m} for H3.6M, Fig. \ref{fig:x3dhp} for 3DHP, and Fig. \ref{fig:x3dpw} for 3DPW. Fig. \ref{fig:xh36m} and Fig. \ref{fig:x3dhp} correspond to the tasks discussed in the alternative order in Sec. \ref{sec:quant}. The qualitative comparisons involve \textbf{Source-only}, \textbf{CoTTA-Pose} \cite{wang2022continual}, \textbf{PoseDA-LL} \cite{chai2023poseda}, \textbf{Ours}, and \textbf{Ground Truth}. Visual inspection reveals a significant performance advantage of our method over other baselines. These qualitative results in Fig. \ref{fig:xh36m} and Fig. \ref{fig:x3dhp} also highlight the effectiveness and robustness of our proposed method in various data arrival scenarios.

\section{Extra Ablation Study}
\label{sec:ab}

\begin{table}[!htb]
\caption{Ablation study on the task H3.6M: S1 $\rightarrow$ S5, S6, S7, S8 when arriving at the final timestamp $t = 4$.}
  \begin{subtable}{0.52\linewidth}
  \caption{3D pose generators.}
  \resizebox{1.0\linewidth}{!}{
  \begin{tabular}{ c | c  c  c  c  c}
    \hline
      Method &  S5 & S6 & S7 & S8 & Avg \\
    \hline
     {Ours w/o PS} & 49.2/43.3 & 49.6/41.7 & 45.0/39.1 & 42.8/29.2 & 46.7/38.3\\
     {Ours w/o TE} & 49.5/43.4 & 49.9/41.9 & 44.7/39.0 & 43.3/29.4 & 46.9/38.4\\
     {Ours w/o DE} & 50.8/44.2 & 50.8/42.4 & 44.3/38.4 & 40.5/28.1 & 46.6/38.3\\
     {Ours} & \textbf{48.7/42.5} & \textbf{48.6/40.8} & \textbf{42.3/36.9} & \textbf{40.0/27.4} & \textbf{44.9/36.9} \\

    \bottomrule
  \end{tabular}}
  \end{subtable}
  \begin{subtable}{0.44\linewidth}
  \caption{Overall framework.}
  \resizebox{1.0\linewidth}{!}{
  \begin{tabular}{ c | c  c  c  c  c}
    \hline
      Method &  S5 & S6 & S7 & S8 & Avg \\
    \hline
     {Ours w/o $\mathcal{L}_{2D}$} & 49.3/43.0 & 49.1/41.4 & 43.8/37.7 & 42.5/28.8 & 46.2/37.7\\
     {Ours w/o $\mathcal{L}_{3D}$} & 49.6/43.3 & 49.8/41.9 & 43.6/37.7 & 43.4/29.1 & 46.6/38.0 \\
     {Ours w/o $\mathcal{L}_{dis}$} & 51.3/44.5 & 51.4/42.6 & 45.0/38.9 & 43.9/29.3 & 47.9/38.8\\
     {Ours w/o EMA} & 50.9/44.2 & 51.0/42.5 & 42.9/37.1 & 40.6/28.2 & 46.1/37.8\\
     {Ours} & \textbf{48.7/42.5} & \textbf{48.6/40.8} & \textbf{42.3/36.9} & \textbf{40.0/27.4} & \textbf{44.9/36.9} \\

    \bottomrule
    \end{tabular}}
  \end{subtable}
\label{tab:ab-h36m}
\end{table}

\begin{table}[!htb]
\caption{Ablation study on the task H3.6M $\rightarrow$ 3DHP: TS1,TS2,...,TS6 when arriving at the final timestamp $t = 6$.}
  \begin{subtable}{0.52\linewidth}
  \caption{3D pose generators.}
  \resizebox{1.0\linewidth}{!}{
  \begin{tabular}{ c | c  c  c  c  c  c  c}
    \hline
      Method &  TS1 & TS2 & TS3 & TS4 & TS5 & TS6 & Avg  \\
    \hline
     {Ours w/o PS} & 63.7/45.8 & 78.3/52.6 & 65.2/44.0 & 77.5/53.9 & 86.5/58.3 & 97.8/64.4 & 78.0/53.2\\
     {Ours w/o TE} & 63.1/45.4 & 77.6/52.0 & 64.4/43.3 & 77.0/53.5 & 85.1/57.9 & 96.9/63.7 & 77.4/52.7 \\
     {Ours w/o DE} & 65.2/46.6 & 79.4/53.2 & 66.7/44.5 & 79.7/55.7 & 84.3/57.2 & 95.5/63.1 & 78.6/53.4\\
     {Ours} & \textbf{61.1/42.9} & \textbf{74.9/49.7} & \textbf{62.3/40.9} & \textbf{75.4/52.3} & \textbf{83.9/55.5} & \textbf{94.3/62.6} & \textbf{75.3/50.7} \\
    \bottomrule
  \end{tabular}}
  \end{subtable}
  \hfill
  \begin{subtable}{0.44\linewidth}
  \caption{Overall framework.}
  \resizebox{1.0\linewidth}{!}{
  \begin{tabular}{ c | c  c  c  c  c  c  c}
    \hline
      Method & TS1 & TS2 & TS3 & TS4 & TS5 & TS6 & Avg  \\
    \hline
     {Ours w/o $\mathcal{L}_{2D}$} & 64.5/46.3 & 78.8/52.7 & 66.0/44.2 & 79.1/55.2 & 84.6/55.8 & 95.1/63.0 & 78.0/52.9\\
     {Ours w/o $\mathcal{L}_{3D}$} & 65.9/47.5 & 80.3/53.7 & 67.3/45.3 & 80.6/56.2 & 86.0/56.7 & 96.6/64.0 & 79.5/53.9\\
     {Ours w/o $\mathcal{L}_{dis}$} & 67.0/47.9 & 81.2/54.3 & 68.5/45.7 & 81.7/56.9 & 87.1/57.1 & 97.8/64.7 & 80.5/54.4\\
     {Ours w/o EMA} & 63.0/43.5 & 76.5/50.9 & 63.7/42.2 & 76.8/52.7 & 86.0/56.8 & 97.1/64.7 & 77.2/51.8\\
     {Ours} & \textbf{61.1/42.9} & \textbf{74.9/49.7} & \textbf{62.3/40.9} & \textbf{75.4/52.3} & \textbf{83.9/55.5} & \textbf{94.3/62.6} & \textbf{75.3/50.7} \\

    \bottomrule
  \end{tabular}}
  \end{subtable}
\label{tab:ab-hp3d}
\end{table}

In this section, we present additional ablation studies on the modules of the 3D pose generators and the overall framework based on the other two tasks, as shown in Tab. \ref{tab:ab-h36m} and Tab. \ref{tab:ab-hp3d}. The results support our conclusion in the main paper that all these modules and loss functions are essential for the lifelong domain adaptation in the 3D HPE method.

\section{Analysis of 2D Poses}
\label{sec:ab-2d}

\begin{table}[tb]
\caption{Analysis of using (a) DET \cite{detectron2018} and (b) HRNet \cite{wang2020deep} to generate 2D poses when $t = 2$ for the two multi-dataset adaptation tasks  "H3.6M$\rightarrow$3DHP$\rightarrow$3DPW" and "H3.6M$\rightarrow$3DPW$\rightarrow$3DHP". }
  \begin{subtable}{0.49\linewidth}
  \caption{Results of using DET \cite{detectron2018} to generate 2D poses.}
  \resizebox{1.0\linewidth}{!}{
  \begin{tabular}{ c | c  c | c  c}
    \hline
      & \multicolumn{2}{c|}{H3.6M $\rightarrow$ 3DHP $\rightarrow$ 3DPW} &
      \multicolumn{2}{c}{H3.6M $\rightarrow$ 3DPW $\rightarrow$ 3DHP} \\
      \hline
       Method & 3DHP & 3DPW & 3DHP & 3DPW \\
    \hline
    {Adaptpose-LL} & 94.2/67.9 & 91.8/54.3 & 83.8/56.4 & 99.8/66.1 \\
    {RMT-Pose} & 94.9/67.8 & 92.4/54.9 & 83.3/56.2 & 98.7/63.5 \\
    {CoTTA-Pose} & 93.7/66.5 & 92.4/54.8 & 83.6/56.3 & 97.1/62.1 \\
    {PoseDA-LL} & 92.5/66.0 & 91.2/53.6 & 83.3/55.8 & 96.6/57.4 \\
    {Ours} & \textbf{79.3/54.8} & \textbf{85.5/49.2} & \textbf{82.1/55.0} & \textbf{87.1/51.3} \\

    \bottomrule
  \end{tabular}}
    \label{tab:ab-det}
  \end{subtable}
  \hfill
  \begin{subtable}{0.49\linewidth}
  \caption{Results of using HRNet \cite{wang2020deep} to generate 2D poses.}
  \resizebox{1.0\linewidth}{!}{
  \begin{tabular}{ c | c  c | c  c}
    \hline
      & \multicolumn{2}{c|}{H3.6M $\rightarrow$ 3DHP $\rightarrow$ 3DPW} &
      \multicolumn{2}{c}{H3.6M $\rightarrow$ 3DPW $\rightarrow$ 3DHP} \\
      \hline
      
     Method & 3DHP & 3DPW & 3DHP & 3DPW \\
    \hline
    {Adaptpose-LL} & 93.1/66.8 & 90.7/53.2 & 82.9/55.8 & 99.2/65.3 \\
    {RMT-Pose} & 93.7/66.7 & 91.2/53.9 & 82.4/55.6 & 97.6/62.9 \\
    {CoTTA-Pose} & 92.5/65.4 & 90.9/53.8 & 82.6/56.1 & 95.9/61.6 \\
    {PoseDA-LL} & 91.3/64.8 & 90.0/52.4 & 82.2/55.4 & 94.5/56.2 \\
    {Ours} & \textbf{78.2/52.7} & \textbf{83.6/47.1} & \textbf{80.1/53.7} & \textbf{85.2/48.4} \\

    \bottomrule
  \end{tabular}}
  \label{tab:ab-hr}
  \end{subtable}
\label{tab:ab-2D}
\end{table}

In this section, we investigate the impact of 2D predictions on lifelong domain adaptive 3D HPE. Following established protocols from prior works \cite{gholami2022adaptpose,chai2023poseda,liu2023posynda}, the main paper utilizes ground truth 2D poses as input. In Tab. \ref{tab:ab-2D}, we explore the use of two widely-used 2D pose estimators to generate 2D poses. Results based on DET \cite{detectron2018} are presented in Tab. \ref{tab:ab-det}, while those based on HRNet \cite{wang2020deep} are shown in Tab. \ref{tab:ab-hr}.

Upon analyzing these results, we find that our proposed method consistently outperforms PoseDA-LL significantly under both settings. For example, when utilizing DET to generate 2D poses, our approach maintains a lead of 13.2mm for MPJPE and 11.2mm for PA-MPJPE on 3DHP for the task "H3.6M$\rightarrow$3DHP$\rightarrow$3DPW". Although there is a certain level of degradation compared to results using ground truth 2D poses, the performance remains noteworthy. The results demonstrate the effectiveness and robustness of our method when utilizing distinctive 2D poses as the input.

\section{Analysis of 2D-lifting-3D Backbones}
\label{sec:ab-3d}

\begin{table}[tb]
\caption{Analysis of using (a) PoseFormer \cite{zheng20213d} and (b) MixSTE \cite{zhang2022mixste} as the 2D-lifting-3D backbones when $t = 2$ for the two tasks  "H3.6M$\rightarrow$3DHP$\rightarrow$3DPW" and "H3.6M$\rightarrow$3DPW$\rightarrow$3DHP". }
  \begin{subtable}{0.49\linewidth}
  \caption{Results of using PoseFormer \cite{zheng20213d} as the 2D-lifting-3D backbone.}
  \resizebox{1.0\linewidth}{!}{
  \begin{tabular}{ c | c  c | c  c}
    \hline
      & \multicolumn{2}{c|}{H3.6M $\rightarrow$ 3DHP $\rightarrow$ 3DPW} &
      \multicolumn{2}{c}{H3.6M $\rightarrow$ 3DPW $\rightarrow$ 3DHP} \\
      \hline
       Method & 3DHP & 3DPW & 3DHP & 3DPW \\
    \hline
    {Adaptpose-LL} & 89.6/63.3 & 87.2/49.6 & 79.5/52.6 & 96.7/62.1 \\
    {RMT-Pose} & 89.0/63.1 & 87.3/49.8 & 79.3/52.3 & 94.5/61.1 \\
    {CoTTA-Pose} & 89.9/63.6 & 87.2/49.7 & 80.1/52.4 & 92.3/59.7 \\
    {PoseDA-LL} & 88.0/61.3 & 86.7/48.4 & 79.0/52.0 & 91.1/53.1 \\
    {Ours} & \textbf{74.6/50.2} & \textbf{80.9/44.7} & \textbf{76.6/51.2} & \textbf{82.4/45.7} \\

    \bottomrule
  \end{tabular}}
    \label{tab:ab-pf}
  \end{subtable}
  \hfill
  \begin{subtable}{0.49\linewidth}
  \caption{Results of using MixSTE \cite{zhang2022mixste} as the 2D-lifting-3D backbone.}
  \resizebox{1.0\linewidth}{!}{
  \begin{tabular}{ c | c  c | c  c}
    \hline
      & \multicolumn{2}{c|}{H3.6M $\rightarrow$ 3DHP $\rightarrow$ 3DPW} &
      \multicolumn{2}{c}{H3.6M $\rightarrow$ 3DPW $\rightarrow$ 3DHP} \\
      \hline
      
     Method & 3DHP & 3DPW & 3DHP & 3DPW \\
    \hline
    {Adaptpose-LL} & 88.7/62.6 & 86.4/49.0 & 78.7/51.8 & 95.8/61.3 \\
    {RMT-Pose} & 88.1/62.4 & 86.2/49.2 & 78.5/51.5 & 93.6/60.6 \\
    {CoTTA-Pose} & 89.0/62.9 & 86.3/49.1 & 79.3/51.5 & 91.8/59.2 \\
    {PoseDA-LL} & 87.1/60.6 & 85.6/48.3 & 78.0/51.1 & 90.8/52.3 \\
    {Ours} & \textbf{73.1/49.7} & \textbf{80.4/44.2} & \textbf{75.0/50.5} & \textbf{82.1/45.2} \\

    \bottomrule
  \end{tabular}}
  \label{tab:ab-mixste}
  \end{subtable}
\label{tab:ab-3d}
\end{table}

In this section, we explore the influence of 2D-to-3D backbone models on lifelong domain adaptive 3D HPE. Following established protocols from previous works \cite{gholami2022adaptpose,chai2023poseda,liu2023posynda}, the main paper utilizes VideoPose3D \cite{pavllo20193d}. In Tab. \ref{tab:ab-3d}, we investigate the performance of two widely-used 2D-to-3D backbone models in predicting 3D poses. Results obtained using PoseFormer \cite{zheng20213d} are presented in Tab. \ref{tab:ab-pf}, while those utilizing MixSTE \cite{zhang2022mixste} are shown in Tab. \ref{tab:ab-mixste}.

Upon analyzing these values, it becomes evident that our proposed method consistently outperforms PoseDA-LL significantly under both settings. For example, when employing PoseFormer to predict 3D poses, our approach maintains a lead of 13.9mm for MPJPE and 11.4mm for PA-MPJPE on 3DHP for the task "H3.6M$\rightarrow$3DHP$\rightarrow$3DPW". Furthermore, the results obtained using the two backbones PoseFormer and MixSTE surpass those relying on VideoPose3D.

\section{Analysis of the Number of Frames}
\label{sec:ab-frame}

\begin{table}[tb]
\caption{Analysis of (a) 9-frame and (b) 81-frame when $t = 2$ for the two multi-dataset adaptation tasks  "H3.6M$\rightarrow$3DHP$\rightarrow$3DPW" and "H3.6M$\rightarrow$3DPW$\rightarrow$3DHP". }
  \begin{subtable}{0.49\linewidth}
  \caption{Results of 9-frame scenario.}
  \resizebox{1.0\linewidth}{!}{
  \begin{tabular}{ c | c  c | c  c}
    \hline
      & \multicolumn{2}{c|}{H3.6M $\rightarrow$ 3DHP $\rightarrow$ 3DPW} &
      \multicolumn{2}{c}{H3.6M $\rightarrow$ 3DPW $\rightarrow$ 3DHP} \\
      \hline
       Method & 3DHP & 3DPW & 3DHP & 3DPW \\
    \hline
    {Adaptpose-LL} & 92.1/65.3 & 89.4/51.6 & 82.0/54.8 & 99.0/64.3 \\
    {RMT-Pose} & 92.5/65.5 & 89.8/51.8 & 81.4/54.7 & 97.0/63.5 \\
    {CoTTA-Pose} & 91.8/65.0 & 90.0/51.7 & 82.5/54.8 & 95.2/61.5 \\
    {PoseDA-LL} & 90.6/63.7 & 89.3/50.4 & 81.4/54.9 & 93.5/55.3 \\
    {Ours} & \textbf{76.7/52.3} & \textbf{82.7/47.8} & \textbf{79.8/54.4} & \textbf{84.7/48.6} \\

    \bottomrule
  \end{tabular}}
    \label{tab:ab-9f}
  \end{subtable}
  \hfill
  \begin{subtable}{0.49\linewidth}
  \caption{Results of 81-frame scenario.}
  \resizebox{1.0\linewidth}{!}{
  \begin{tabular}{ c | c  c | c  c}
    \hline
      & \multicolumn{2}{c|}{H3.6M $\rightarrow$ 3DHP $\rightarrow$ 3DPW} &
      \multicolumn{2}{c}{H3.6M $\rightarrow$ 3DPW $\rightarrow$ 3DHP} \\
      \hline
      
     Method & 3DHP & 3DPW & 3DHP & 3DPW \\
    \hline
    {Adaptpose-LL} & 89.9/63.4 & 87.3/50.1 & 80.0/52.8 & 97.1/62.3 \\
    {RMT-Pose} & 90.3/63.6 & 87.7/50.3 & 79.4/52.2 & 94.8/61.2 \\
    {CoTTA-Pose} & 89.4/62.9 & 88.0/50.2 & 80.5/52.7 & 92.7/59.8 \\
    {PoseDA-LL} & 88.2/62.0 & 86.9/49.0 & 79.1/52.6 & 91.3/53.2 \\
    {Ours} & \textbf{74.8/50.3} & \textbf{80.2/45.5} & \textbf{75.8/50.9} & \textbf{82.6/45.5} \\

    \bottomrule
  \end{tabular}}
  \label{tab:ab-81f}
  \end{subtable}
\label{tab:ab-frame-sm}
\end{table}

In this section, we explore the impact of varying the number of frames on lifelong domain adaptive 3D HPE. Following established protocols from prior research \cite{gholami2022adaptpose,chai2023poseda,liu2023posynda}, the main paper employs the 27-frame setting. In Tab. \ref{tab:ab-frame-sm}, we scrutinize the performance under the 9-frame setting presented in Tab. \ref{tab:ab-9f}, and the 81-frame scenario detailed in Tab. \ref{tab:ab-81f}.

Upon examination of these outcomes, it is evident that our proposed method consistently outperforms PoseDA-LL significantly across both settings. For instance, in the 81-frame scenario, our approach maintains a lead of 13.4mm for MPJPE and 11.7mm for PA-MPJPE on 3DHP for "H3.6M$\rightarrow$3DHP$\rightarrow$3DPW". Moreover, while the 27-frame configuration outperforms the 9-frame setting, it falls short of the 81-frame setup, suggesting that a greater number of frames can indeed contribute to enhanced performance.

\section{Analysis of Domain-aware Priors' Generation}
\label{sec:ab-gen}

In this section, we conduct a comparative analysis of different methods for generating domain-aware priors. In the main paper, DDIM \cite{song2020denoising} is employed to train the 2D pose sampler. Here, we choose three representative generation methods for comparison: no priors, GAN \cite{goodfellow2014generative}, and DDPM \cite{ho2020denoising}. The results of these comparisons are presented in Tab. \ref{tab:ab-priors}.

\begin{table}
\centering
  \caption{Analysis of the domain-aware priors' generation methods when $t = 2$ for the two multi-dataset adaptation tasks  "H3.6M$\rightarrow$3DHP$\rightarrow$3DPW" and "H3.6M$\rightarrow$3DPW$\rightarrow$3DHP". }
  \resizebox{1.0\linewidth}{!}{
  \begin{tabular}{c | c  c | c  c}
    \hline
      & \multicolumn{2}{c|}{H3.6M $\rightarrow$ 3DHP $\rightarrow$ 3DPW} &
      \multicolumn{2}{c}{H3.6M $\rightarrow$ 3DPW $\rightarrow$ 3DHP} \\
      \hline
      
    Generation Methods & 3DHP & 3DPW & 3DHP & 3DPW \\
    \hline

     {No Priors} & 83.5/57.4 & 83.7/47.6 & 79.2/52.7 & 88.2/51.7\\ 
     {GAN \cite{goodfellow2014generative}} & 81.2/56.9 & 83.1/47.0 & 77.9/51.2 & 86.9/50.5 \\
     {DDPM \cite{ho2020denoising}} & 76.7/51.5 & 82.9/46.3 & 78.7/52.6 & 85.1/48.8 \\
     {DDIM \cite{song2020denoising} (Ours)} & \textbf{75.3/51.1} & \textbf{81.7/45.6} & \textbf{78.3/52.2} & \textbf{83.7/46.9}\\

    \bottomrule
  \end{tabular}}\vspace{-10pt}
  \label{tab:ab-priors}
\end{table}

Analyzing the values in Tab. \ref{tab:ab-priors}, it becomes evident that domain-aware priors play a pivotal role in mitigating catastrophic forgetting on previous data. Additionally, diffusion-based priors demonstrate a significant performance advantage over the GAN-based priors, highlighting the efficacy of diffusion models in preserving mode coverage and diversity while alleviating mode collapse. Furthermore, our choice of DDIM outperforms the vanilla diffusion model DDPM.

\section{Parameter Analysis}
\label{sec:param}

In this section, we delve into a comprehensive parameter analysis, building upon the optimization process outlined in the main paper. Three key hyperparameters, $\alpha$, $\beta$, and $\gamma$, are introduced for this analysis. Given the symmetry between Eq. 4 and Eq. 5, we consolidate $\beta$ and $\gamma$ into a single value during the tuning process. The outcomes associated with $\alpha$ are presented in Tab. \ref{tab:ab-alphas}, while those pertaining to $\beta$ and $\gamma$ are detailed in Tab. \ref{tab:ab-beta-gamma}.

\begin{table}
\centering
  \caption{Analysis of $\alpha$ values when $t = 2$ for the two multi-dataset adaptation tasks  "H3.6M$\rightarrow$3DHP$\rightarrow$3DPW" and "H3.6M$\rightarrow$3DPW$\rightarrow$3DHP". }
  \resizebox{1.0\linewidth}{!}{
  \begin{tabular}{c | c  c | c  c}
    \hline
      & \multicolumn{2}{c|}{H3.6M $\rightarrow$ 3DHP $\rightarrow$ 3DPW} &
      \multicolumn{2}{c}{H3.6M $\rightarrow$ 3DPW $\rightarrow$ 3DHP} \\
      \hline
      
    Values of $\alpha$ & 3DHP & 3DPW & 3DHP & 3DPW \\
    \hline

     {$\alpha = 0.20$} & 76.9/52.0 & 82.8/47.4 & 79.5/52.7 & 85.1/47.8 \\
     {$\alpha = 0.25$} & 76.0/51.6 & 82.3/46.7 & 78.8/52.4 & 84.2/47.1 \\
     {$\alpha = 0.30$} & 75.6/51.4 & 82.0/\textbf{45.6} & 78.5/\textbf{52.2} & 83.8/47.0 \\
     {$\alpha = 0.35$ (Ours)} & \textbf{75.3/51.1} & \textbf{81.7/45.6} & \textbf{78.3/52.2} & \textbf{83.7/46.9}\\
     {$\alpha = 0.40$} & 75.7/51.4 & 81.9/45.7 & 78.7/52.3 & 84.0/47.2 \\ 
     {$\alpha = 0.45$} & 76.3/51.9 & 82.6/46.9 & 78.8/52.5 & 84.6/47.3 \\
     {$\alpha = 0.50$} & 77.2/52.4 & 83.2/47.8 & 79.9/52.9 & 85.3/48.1 \\

    \bottomrule
  \end{tabular}}\vspace{-10pt}
  \label{tab:ab-alphas}
\end{table}

\begin{table}
\centering
  \caption{Analysis of $\beta$ and $\gamma$ values when $t = 2$ for the two multi-dataset adaptation tasks  "H3.6M$\rightarrow$3DHP$\rightarrow$3DPW" and "H3.6M$\rightarrow$3DPW$\rightarrow$3DHP". }
  \resizebox{1.0\linewidth}{!}{
  \begin{tabular}{c | c  c | c  c}
    \hline
      & \multicolumn{2}{c|}{H3.6M $\rightarrow$ 3DHP $\rightarrow$ 3DPW} &
      \multicolumn{2}{c}{H3.6M $\rightarrow$ 3DPW $\rightarrow$ 3DHP} \\
      \hline
      
    Values of $\beta$ and $\gamma$ & 3DHP & 3DPW & 3DHP & 3DPW \\
    \hline

     { 1.0 } & 76.8/52.3 & 83.4/46.4 & 79.5/52.7 & 84.5/47.8\\ 
     { 1.5 } & 76.0/51.8 & 82.6/46.2 & 79.4/52.7 & 84.3/47.5 \\
     { 2.0 } & 75.6/51.4 & 82.2/45.7 & 78.9/52.4 & 83.8/47.1 \\
     {2.5 (Ours)} & \textbf{75.3/51.1} & \textbf{81.7/45.6} & \textbf{78.3/52.2} & \textbf{83.7/46.9}\\
     { 3.0 } & 75.8/51.5 & 81.9/45.5 & 78.5/52.3 & 84.0/47.2 \\ 
     { 3.5 } & 76.6/52.2 & 82.9/46.1 & 79.0/52.7 & 84.6/47.9 \\
     { 4.0 } & 77.2/52.4 & 83.3/46.4 & 79.8/52.9 & 85.0/48.3 \\

    \bottomrule
  \end{tabular}}\vspace{-10pt}
  \label{tab:ab-beta-gamma}
\end{table}

According to these results, we note that our choices of hyperparameters are reasonable. Additionally, our proposed method exhibits robustness, as it displays low sensitivity to changes in hyperparameters.

\section{Details of Part Segments}
\label{sec:part}

\begin{figure*}[!htb]
  \centering
  \includegraphics[width=1.0\linewidth]{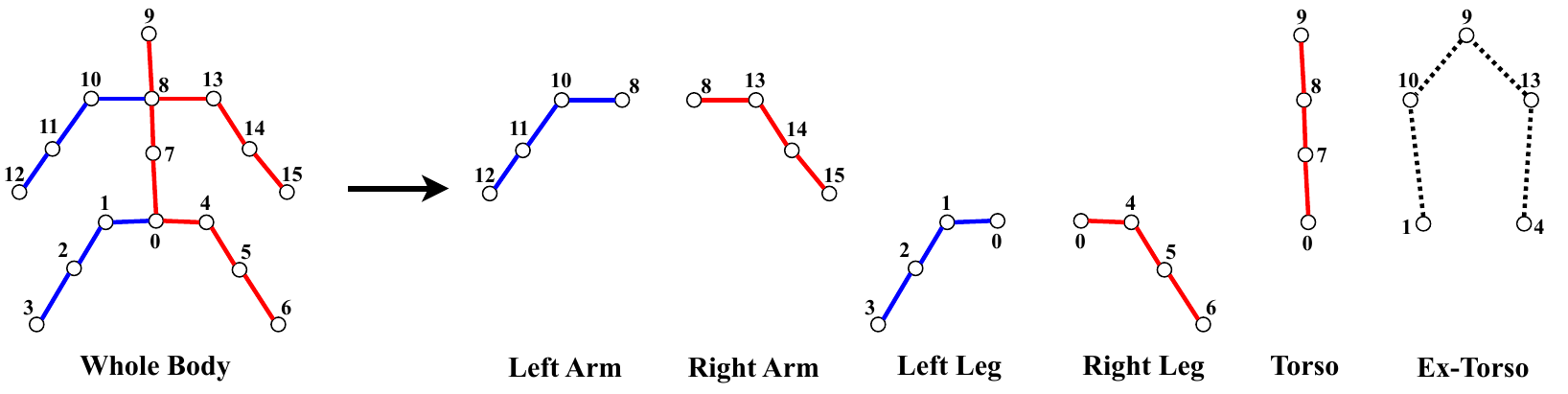}
  \caption{Details of the part segments we used for 3D pose generators.}
  \label{fig:parts}
\end{figure*}

In this section, we discuss the details of the body part segments that we applied in encoding the 3D pose generators, as shown in Fig. \ref{fig:parts}.

Beyond the traditional five segments—left arm, right arm, left leg, right leg, and the torso—commonly employed in many studies on human pose and shape, we introduce an additional segment, ex-torso \cite{akhter2015pose}, to establish connections between joints that are not explicitly linked but still share relationships. This approach enables the learned pose-aware representations to better capture the complexities of the human pose.

\section{Explanation of the Three Generators}
\label{sec:exp}

There are three 3D generators (unified paradigm in Fig. 3 of the main paper) in the framework (Fig. 2 of the main paper) to augment 3D pose. The bone angle generator $G_{BA}$ generates novel unit vectors for all the bones. The bone length generator $G_{BL}$ changes the lengths of bones in a ratio between $-30\%$ and $30\%$. The rotation and translation generator $G_{RT}$ synthesizes new quarternion and translation value for the rotation and translation operation.

\section{Details of Domain-aware Encoding}
\label{sec:diffusion}

In this section, we provide details of domain-aware encoding, which includes two parts. Assume now we focus on domain $D_t$, where $t$ is the current timestamp. First is the preparation of the diffusion sampler. This process is in parallel with all the previous target domains' adaptation. For any target domain $D_\tau$ ($1 \leq \tau < t$), its pretraining is shown in Alg. \ref{alg:pretrain}. Second is the sampling on the current domain. This process is a part of the current domain's adaptation, which is exhibited in Alg. \ref{alg:sample}.

\begin{algorithm}
\caption{Pretrain Diffusion Sampler on a Previous Domain $\mathcal{D}_{\tau}$ before timestamp $t$}
\label{alg:pretrain}
\begin{algorithmic}[1]
\Require 2D poses domain $\mathcal{D}_{\tau}$ before timestamp $t$ ($1 \leq \tau < t$) ,  maximum diffusion steps $T$, pretrained diffusion sampler $\theta_{\tau - 1}$ before timestamp $\tau$

\State Initialize 2D pose diffusion sampler $\theta_{\tau}$ with $\theta_{\tau - 1}$.
\For{each training iteration}
    \State Sample 2D poses $\mathbf{x}_0 \sim \mathcal{D}_{\tau}$
    \State Sample training step $k \sim \text{Uniform}(1, T)$ 
    \State Sample noise $\boldsymbol{\epsilon} \sim \mathcal{N}(0, \mathbf{I})$ 
    \State Compute noisy poses $\mathbf{x}_k = \sqrt{\alpha_k}\mathbf{x}_0 + \sqrt{1-\alpha_k}\boldsymbol{\epsilon}$ \Comment{Add noise to the poses}
    \State Optimize $\theta_{\tau - 1}$ by minimizing $\|\boldsymbol{\epsilon} - \boldsymbol{\epsilon}_{\theta_{\tau - 1}}(\mathbf{x}_k, k)\|^2$
\EndFor
\State \Return 2D pose diffusion sampler $\theta_{\tau}$

\end{algorithmic}
\end{algorithm}

\begin{algorithm}
\caption{Generate Domain-aware Priors via DDIM Sampling}
\label{alg:sample}
\begin{algorithmic}[1]
\Require Current domain data $\mathcal{D}_t$ with $N$ samples, maximum diffusion steps $T' = T / 10$, pretrained diffusion sampler $\theta_{t-1}$ before timestamp $t$, DDIM sampling step size $\eta = 0.2$ 
\State Initialize domain-aware prior set $\mathcal{P} = \emptyset$
\For{$i = 1$ to $N$} 
    \State Sample $\mathbf{x}_{T'} \sim \mathcal{N}(0, \mathbf{I})$ 
    \For{$k = T'$ down to $1$} 
        \State Predict noise $\boldsymbol{\epsilon}_{\theta_{t-1}}(\mathbf{x}_k, k)$ 
        \State Predict $\mathbf{x}_0$ from $\mathbf{x}_k$: $\hat{\mathbf{x}}_0 = \frac{\mathbf{x}_k - \sqrt{1-\alpha_k}\boldsymbol{\epsilon}_{\theta_{t-1}}(\mathbf{x}_k, k)}{\sqrt{\alpha_k}}$
        \State $\mathbf{x}_{k-1} = \sqrt{\alpha_{k-1}}\hat{\mathbf{x}}_0 + \sqrt{1 - \alpha_{k-1} - \sigma_k^2}\boldsymbol{\epsilon}_{\theta_{t-1}}(\mathbf{x}_k, k) + \sigma_k\boldsymbol{\epsilon}$ where $\boldsymbol{\epsilon} \sim \mathcal{N}(0, \mathbf{I})$ if $\eta > 0$, otherwise $\boldsymbol{\epsilon} = \mathbf{0}$
        \State where $\sigma_k = \eta\sqrt{\frac{1-\alpha_{k-1}}{1-\alpha_k}}\sqrt{1 - \frac{\alpha_k}{\alpha_{k-1}}}$
    \EndFor
    \State Add generated 2D pose: $\mathcal{P} = \mathcal{P} \cup \{\mathbf{x}_0\}$
\EndFor
\State \Return Domain-aware prior set $\mathcal{P}$
\end{algorithmic}
\end{algorithm}

\section{Comparisons of Different Encoding Emthods}

\label{sec:encode}

The key distinction lies in how the generators encode their inputs. Prior works \cite{chai2023poseda, gholami2022adaptpose} rely on joint coordinates or bone vectors for pose-aware encoding. In contrast, our approach introduces part segments for pose-aware encoding, enhanced by temporal-aware and domain-aware components. Our comprehensive encoding scheme demonstrates superior performance, as shown by the comparative results in Tab. \ref{tab:encode}.

\begin{table*}[!ht]
    \centering
    \caption{Comparisons of different encoding methods.}\vspace{-10pt}
    \resizebox{1.0\linewidth}{!}{
    \begin{tabular}{c  c  c  c  c  c  c  c  c   c}
         \toprule
          Task & Time & Encoding Method & S5 & S6 & S7 & S8 & - & - &  Avg\\

         \hline
         \multirow{2}{*}{S1 $\rightarrow$S5,S6,S7,S8} & \multirow{2}{*}{t = 4} & {\cite{chai2023poseda, gholami2022adaptpose} } & 52.4/45.6 & 53.8/45.1 & 47.0/40.7 & 41.9/30.0 & - & - & 48.8/40.4\\
         & & {Ours} & \textbf{48.7/42.5} & \textbf{48.6/40.8} & \textbf{42.3/36.9} & \textbf{40.0/27.4} & - & - &  \textbf{44.9/36.9} \\
         \hline
         Task & Time & Encoding Method & TS1 & TS2 & TS3 & TS4 & TS5 & TS6 &  Avg\\
         \hline
         \multirow{2}{*}{H3.6M $\rightarrow$TS1,..,TS6} & \multirow{2}{*}{t = 6} & {\cite{chai2023poseda, gholami2022adaptpose} } & 67.5/48.3 & 83.0/54.7 & 68.3/45.2 & 81.4/56.6 & 89.0/58.5 & 95.8/65.3 & 80.8/54.8\\
         && {Ours} & \textbf{61.1/42.9} & \textbf{74.9/49.7} & \textbf{62.3/40.9} & \textbf{75.4/52.3} & \textbf{83.9/55.5} & \textbf{94.3/62.6} & \textbf{75.3/50.7}\\

         \bottomrule
    \end{tabular}}%
\label{tab:encode}
\end{table*}

\section{Computational Complexity and Runtime Analysis}
\label{sec:computational_analysis}

In this section, we provide a detailed analysis of the computational overhead introduced by our proposed Diffusion Encoding (DE) component and compare the training times across different method configurations. Tab. \ref{tab:time_comparison} presents a comprehensive comparison of training times for our method against the baseline PoseDA-LL approach on the H3.6M $\rightarrow$ TS1,..,TS6 domain adaptation task. The training process consists of three distinct phases: pretraining, diffusion encoding, and domain adaptation.

\begin{table}[!ht]
\centering
\caption{Training time comparison on H3.6M $\rightarrow$ TS1,..,TS6 domain adaptation task}
\resizebox{1.0\linewidth}{!}{%
\begin{tabular}{c|ccc}
\toprule
Training Phase & PoseDA-LL & Ours w/o DE & Ours (Full) \\
\hline
Pretrain Epoch/Time & 40/346min & 40/346min & 40/346min\\
Diffusion Epoch/Time & 0/0min & 0/0min & 10/47min \\
Adaptation Epoch/Time & 30/244min & 30/208min & 30/217min \\
\hline
Total Training Time & 590min & 554min & 610min \\
\hline
MPJPE (mm) & 80.7 & 78.6 & \textbf{75.3} \\
PA-MPJPE (mm) & 54.5 & 53.4 & \textbf{50.7} \\
\bottomrule
\end{tabular}}
\label{tab:time_comparison}
\end{table}


The results demonstrate that our Diffusion Encoding component introduces minimal computational overhead while delivering significant performance improvements. The addition of the diffusion encoding phase requires only 10 epochs (47 minutes) of additional training time, representing less than 8$\%$ increase in total training time compared to the baseline method. Despite this modest computational cost, our full method achieves substantial improvements in both MPJPE (75.3mm vs. 80.7mm, a 6.7$\%$ reduction) and PA-MPJPE (50.7mm vs. 54.5mm, a 7.0$\%$ reduction) compared to PoseDA-LL. Interestingly, our method without diffusion encoding (Ours w/o DE) shows improved training efficiency in the adaptation phase (208min vs. 244min), suggesting that our architectural improvements beyond diffusion encoding also contribute to computational efficiency. The 47-minute investment in diffusion encoding training yields a performance improvement that significantly outweighs the additional computational cost, making our approach highly practical for real-world applications and confirming that our proposed Diffusion Encoding provides an excellent balance between computational efficiency and performance enhancement.

\section{Broader Impacts}
\label{sec:imp}


The introduction of the lifelong domain adaptation setting significantly enhances the real-world applicability of 3D human pose estimation models. By addressing the challenges posed by non-stationary target pose datasets, the proposed framework ensures that pose estimators can effectively adapt to dynamic and evolving scenarios, making them more valuable for real-world applications in diverse fields such as healthcare, sports analysis, and human-computer interaction.

\end{document}